\documentclass[lettersize,journal]{IEEEtran}
\usepackage{amsmath,amsfonts}
\usepackage{algorithmic}
\usepackage{array}
\usepackage[caption=false,font=normalsize,labelfont=sf,textfont=sf]{subfig}
\usepackage{textcomp}
\usepackage{stfloats}
\usepackage{url}
\usepackage{verbatim}
\usepackage{graphicx}
\usepackage[numbers,sort&compress]{natbib}
\usepackage{stmaryrd}
\usepackage{bm}
\usepackage{amssymb}
\usepackage{booktabs}
\usepackage{graphicx}
\usepackage{float}
\usepackage{graphicx}
\usepackage{stfloats}
\usepackage{epstopdf}
\usepackage{algorithm,algorithmic}
\usepackage{multirow}
\usepackage{bbm}
\usepackage{framed}
\usepackage[colorlinks,linkcolor=red,anchorcolor=green,citecolor=green]{hyperref}
\usepackage[numbers,sort&compress]{natbib}
\usepackage{amsthm}
\def\BibTeX{{\rm B\kern-.05em{\sc i\kern-.025em b}\kern-.08em
    T\kern-.1667em\lower.7ex\hbox{E}\kern-.125emX}}
\usepackage{balance}

\DeclareMathOperator*{\argmin}{arg\,min}

\newcommand{\ie}{{\em i.e.}}
\newcommand{\eg}{{\em e.g.}}

\begin{document}
\title{Triply Laplacian Scale Mixture Modeling for Seismic Data Noise Suppression}
\author{Sirui~Pan, Zhiyuan~Zha,~\IEEEmembership{Senior Member,~IEEE},  Shigang~Wang,~\IEEEmembership{Member,~IEEE},\\
 Yue~Li,~\IEEEmembership{Senior Member,~IEEE}, Zipei~Fan,~\IEEEmembership{Member,~IEEE}, Gang~Yan,~\IEEEmembership{Member,~IEEE},\\
 Binh~T.~Nguyen,~\IEEEmembership{Member,~IEEE}, Bihan Wen,~\IEEEmembership{Senior Member,~IEEE}, and Ce Zhu,~\IEEEmembership{Fellow,~IEEE}

\IEEEcompsocitemizethanks{
\IEEEcompsocthanksitem Sirui Pan, Zhiyuan Zha, Shigang Wang, and Yue Li are with the College of Communication Engineering, Jilin University, Changchun 130012, China (Email: pansr24@mails.jlu.edu.cn; zhiyuan\_zha@jlu.edu.cn; wangsg@jlu.edu.cn; liyue@jlu.edu.cn).
\IEEEcompsocthanksitem Zipei Fan is with the School of Artificial Intelligence, Jilin University, Changchun 130012, China. Email: fanzipei@jlu.edu.cn.
\IEEEcompsocthanksitem Gang Yan is with the College of Computer Science and Technology, Jilin University, Changchun 130012, China. Email: gyan8@jlu.edu.cn.
\IEEEcompsocthanksitem Binh T. Nguyen is with the Department of Computer Science, Faculty of Mathematics and Computer Science, University of Science, Vietnam National University Ho Chi Minh City, Ho Chi Minh City 700000, Vietnam. Email: ngtbinh@hcmus.edu.vn.
\IEEEcompsocthanksitem Bihan Wen is with the School of Electrical and Electronic Engineering, Nanyang Technological University, Singapore 639798. Email: bihan.wen@ntu.edu.sg.
\IEEEcompsocthanksitem Ce Zhu is with the Glasgow College, University of Electronic Science and Technology of China, Chengdu 611731, China. Email: eczhu@uestc.edu.cn.
}
}

\markboth{Journal of \LaTeX\ Class Files,~2025}
{Shell \MakeLowercase{\textit{et al.}}: Bare Demo of IEEEtran.cls for IEEE Journals}

\maketitle

\begin{abstract}
Sparsity-based tensor recovery methods have shown great potential in suppressing seismic data noise. These methods exploit tensor sparsity measures capturing the low-dimensional structures inherent in seismic data tensors to remove noise by applying sparsity constraints through soft-thresholding or hard-thresholding operators. However, in these methods, considering that real seismic data are non-stationary and affected by noise, the variances of tensor coefficients are unknown and may be difficult to accurately estimate from the degraded seismic data, leading to undesirable noise suppression performance. In this paper, we propose a novel triply Laplacian scale mixture (TLSM) approach for seismic data noise suppression, which significantly improves the estimation accuracy of both the sparse tensor coefficients and hidden scalar parameters. To make the optimization problem manageable, an alternating direction method of multipliers (ADMM) algorithm is employed to solve the proposed TLSM-based seismic data noise suppression problem. Extensive experimental results on synthetic and field seismic data demonstrate that the proposed TLSM algorithm outperforms many state-of-the-art seismic data noise suppression methods in both quantitative and qualitative evaluations while providing exceptional computational efficiency.
\end{abstract}

\begin{IEEEkeywords}
Seismic data noise suppression, sparsity-based tensor recovery, triply Laplacian scale mixture, ADMM.
\end{IEEEkeywords}

\IEEEpeerreviewmaketitle

\section{Introduction}
\IEEEPARstart{S}{eismic} data provide wealthy geologic information that plays a vital role in various real-world applications, such as petroleum exploration \cite{kuang2021application}, seismic geomorphology \cite{posamentier2022principles}, and earthquake monitoring \cite{mousavi2022deep}. However, due to the influence of acquisition geometry and other factors \cite{marfurt1998suppression,al2004acquisition,falconer2008attribute}, seismic data are inevitably contaminated by a combination of acquisition footprint and Gaussian noise, which not only decreases their visual quality but also hinders their effectiveness in many downstream tasks (\eg, migration \cite{han2022gaussian}, inversion \cite{zhang2021deep}, and stratigraphic imaging \cite{gu2023seismic}). Therefore, developing an effective noise suppression technique is critical for facilitating subsequent seismic data analysis and understanding.

 Over the past decades, various methods have been developed to suppress noise from seismic data. Previous seismic data noise suppression methods are predominantly based on predictive filtering techniques \cite{abma1995lateral,zhang2009footprint,liu2012spatiotemporal,li2017multidimensional}, leveraging the distinctions between signal and noise in the time or frequency domains.  Given an overcomplete dictionary, sparse transform methods represent seismic signals as a linear combination of a limited number of dictionary atoms, which is achieved either through analytic dictionaries with predefined basis functions, such as wavelet \cite{cvetkovic20072d,alali2018attribute}, curvelet \cite{wang2019efficient}, and shearlet transforms \cite{zhang2018multicomponent}, or through learning-based dictionaries adaptively trained on seismic data \cite{gomez2020footprint,liu2021dictionary,zhou2023coherent,chen2021statistics}. Moreover, low-rank (LR) matrix recovery methods have gained widespread application in seismic data noise removal \cite{chen2016simultaneous,oboue2021enhanced,niu2021seismic,lin2023structure,wang2023seismic}, which leverages the assumption that clean seismic data inherently exhibit LR characteristics. However, unfolding multidimensional tensors derived from seismic data into matrix form inevitably destroys their intrinsic structure, leading to suboptimal denoising performance \cite{goldfarb2014robust,chen2018tensor}.

 \begin{figure*}[!t]
\centering
\begin{minipage}[b]{1\linewidth}
{\includegraphics[width= 1\textwidth]{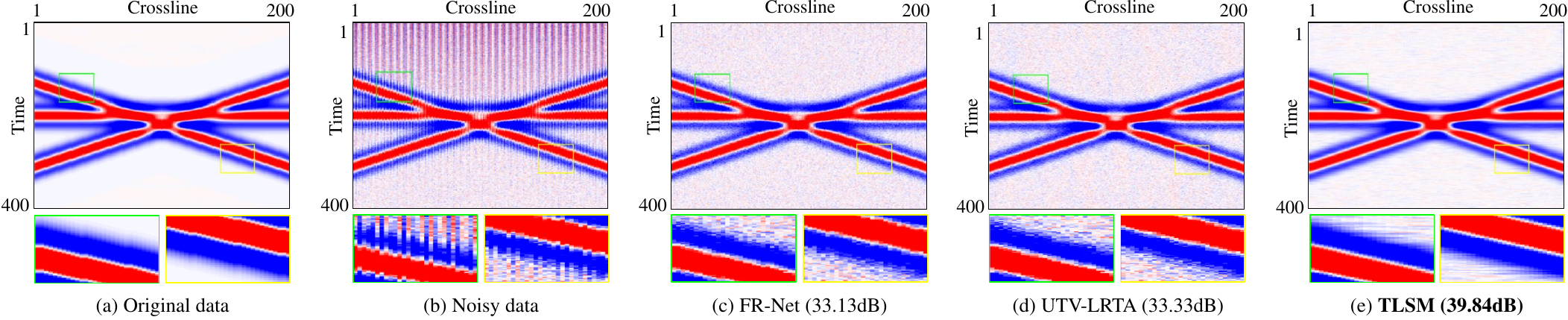}}
\end{minipage}
\vspace{-4mm}
\caption {Comparison of noise suppression results between the proposed TLSM approach and state-of-the-art methods on synthetic data ($F = 0.2, \sigma = 0.03$). (a) Original seismic data; (b) Noisy seismic data; (c) FR-Net \cite{qian2023unsupervised}; (d) UTV-LRTA \cite{qian2023improved}; (e) \textbf{Proposed TLSM}.}
\label{fig:0}
\vspace{-2mm}
\end{figure*}

Recent advancements have shown that sparsity-based tensor recovery methods are effective in suppressing noise in seismic data \cite{feng2021low,feng2021seismic,qian2023improved,liu2024simultaneous}. These methods usually utilize tensor sparsity measures to capture the low-dimensional structures inherent in seismic data tensors, removing noise by applying sparsity penalties through soft-thresholding or hard-thresholding operators. For instance, Feng $\emph{et al}.$ \cite{feng2021low} proposed a seismic data noise suppression method based on an LR tensor minimization model that leverages the spatial similarity and frequency correlation of seismic data. The UTV-LRTA approach \cite{qian2023improved} employed the tensor nuclear norm (TNN) \cite{lu2016tensor} as a measure of tensor sparsity, combined with a unidirectional total variation (UTV) regularizer to acquire the directional characteristics of footprint noise. However, in these sparsity-based tensor recovery methods \cite{feng2021low,feng2021seismic,qian2023improved,liu2024simultaneous}, the non-stationary nature and high susceptibility of seismic data to noise make them difficult to accurately estimate the variances of sparse tensor coefficients from the degraded seismic data, resulting in suboptimal  denoising performance (see an example in Fig.~\ref{fig:0} (d)).

Bearing the above concerns in mind, in this paper, we propose a novel triply Laplacian scale mixture (TLSM) approach for seismic data noise suppression. Unlike most existing methods that enforce sparsity penalties, usually through soft-thresholding or hard-thresholding operators \cite{feng2021low,feng2021seismic,qian2023improved,liu2024simultaneous}, the proposed TLSM approach models each sparse tensor coefficient as a Laplacian distribution with a positive scalar multiplier. By imposing a sparse distribution prior on the scalar multipliers, the proposed TLSM approach enables the joint estimation of both the variances and the values of the sparse tensor coefficients directly from the noisy observations. The
significant contributions of this paper are summarized as follows.

\begin{itemize}
    \item [1)] We propose a sparsity-based tensor recovery framework that incorporates the TLSM priors for seismic data noise suppression.
    \item [2)] To ensure the optimization is both tractable and stable, an alternating direction method of multipliers (ADMM) algorithm is employed to solve the proposed TLSM-based seismic data denoising problem.
    \item [3)] We conduct extensive experiments on both synthetic and field seismic data, demonstrating that the proposed TLSM algorithm not only outperforms many state-of-the-art seismic data denoising methods in both quantitative metrics and visual quality but also achieves superior computational efficiency.

\end{itemize}

The rest of this paper is organized as follows. Section~\ref{sec:2} reviews related works on seismic data noise suppression, encompassing both model optimization and deep learning-based approaches. Section~\ref{sec:3} introduces the relevant notations and essential preliminaries, including the sparsity-based tensor recovery model and LSM modeling. Section~\ref{sec:4} proposes the TLSM model for seismic data noise suppression and develops an efficient algorithm to tackle the associated denoising challenges. Section~\ref{sec:5} presents the experimental results, while Section~\ref{sec:6} concludes the paper.

\section{Related Work}
\label{sec:2}
In this section, we present a brief review of the related works on seismic data noise suppression, with a focus on methods based on model optimization and deep learning.

\subsection{Model Optimization-based Seismic Data Noise Suppression Methods}

Seismic data noise suppression is fundamentally an ill-posed inverse problem \cite{qian2023improved,wang2015improved,li2022simultaneous}. To address this challenge, model optimization-based approaches leverage prior knowledge of seismic data to construct regularization models that refine the solution space, such as total variation (TV) \cite{osher2005iterative}, sparse representation \cite{aharon2006k}, and LR matrix approximation \cite{cai2010singular}. It has been proven that prior-based modeling is of great significance in seismic data noise suppression \cite{chen2021statistics,cheng2015application,chen2016simultaneous,qian2023improved,liu2024simultaneous}. For example, Chen $\emph{et al}.$ \cite{chen2021statistics} proposed a statistics-guided residual dictionary learning (SGRDL) method for seismic data noise removal. Cheng $\emph{et al}.$ \cite{cheng2015application} adopted a nuclear norm constraint under the framework of the LR matrix approximation for seismic data noise suppression. Chen $\emph{et al}.$ \cite{chen2016simultaneous} proposed a damped rank reduction (DRR) approach for seismic data denoising, which employs the block Hankel matrix to decompose the noisy data space into signal and noise subspace. Due to the LR structure embedded in the multidimensional seismic data, LR tensor recovery methods have received considerable attention in seismic data noise suppression \cite{feng2021low,feng2021seismic,qian2023improved,liu2024simultaneous}. For instance, Feng $\emph{et al}.$ \cite{feng2021seismic} employed the CANDECOMP/PARAFAC (CP) decomposition \cite{carroll1970analysis}, integrating a TV constraint for seismic data noise suppression. Qian $\emph{et al}.$ \cite{qian2023improved} proposed a tensor model called UTV-LRTA for seismic data noise suppression, which integrates LR tensor approximation with a unidirectional TV regularizer.

\subsection{Deep Learning-based Seismic Data Noise Suppression Methods}

Deep learning techniques have demonstrated significant potential in seismic data noise suppression \cite{zhu2019seismic,wang2020generative,wang2024efgw,wang2022learning,xu2023deep,qian2023unsupervised,xu2023s2s} by leveraging an end-to-end training approach for deep neural networks (DNNs) \cite{lecun2015deep}. For example, Wang $\emph{et al}.$ \cite{wang2020generative} adopted a generative adversarial network (GAN) \cite{goodfellow2020generative} for seismic denoising, where the generator is used to remove noise and the discriminator guides the generator to restore structural information. The EFGW-UNet method \cite{wang2024efgw} utilized an edge-feature-guided wavelet U-Net \cite{ronneberger2015u} to preserve finer details of effective signals while suppressing noise. However, these supervised learning algorithms \cite{liu2021unsupervised,qiu2021deep} require extensive seismic field data, which is challenging to obtain due to high acquisition costs and limited availability, thereby restricting their applicability. In recent years, various self-supervised learning methods have been developed for seismic data noise suppression \cite{meng2021self,xu2023deep,qian2023unsupervised,xu2023s2s}. For instance, Xu $\emph{et al}.$ \cite{xu2023deep} proposed a deep nonlocal regularizer (DNLR) method for seismic data noise suppression, which combines the learning capability of DNNs with the generalization power of handcrafted regularizers. Qian $\emph{et al}.$ \cite{qian2023unsupervised} introduced a footprint removal network (FR-Net) by regularizing a deep convolutional autoencoder using the UTV. The S2S-WTV method \cite{xu2023s2s} leveraged the Self2Self (S2S) learning framework \cite{quan2020self2self} with a trace-wise masking strategy and weighted TV for seismic data denoising.

\section{Notations and Preliminaries}
\label{sec:3}
\subsection{Notations}
Throughout the paper, the tensor, matrix and vector are denoted as calligraphic letter (\eg, $\cal X$), boldface capital letter (\eg, $\textbf{\emph{X}}$), and bold lowercase letter (\eg, $\textbf{\emph{x}}$), respectively. We define each element ($i_1, i_2,\dots,i_N$) of the tensor ${\cal X}\in{\mathbb R}^{I_1\times I_2\times\dots \times I_N}$ as $\emph{x}_{i_1\times i_2\times\dots \times i_N}$, and ${\cal X}^{(i)}$ corresponds to the ${i}_{th}$ frontal slice of $\cal X$. Each element ($i, j$) of a matrix $\textbf{\emph{X}}$ is represented by $\emph{x}_{i,j}$, and the entry $i$ of a vector $\textbf{\emph{x}}$ is defined as $\emph{x}_i$. The transpose of the matrix $\textbf{\emph{X}}$ and the tensor $\cal X$ are denoted by ${\textbf{\emph{X}}}^T$ and ${\cal X}^T$, respectively. We denote the Frobenius norm of the tensor ${\cal X}$ as $\|{\cal X}\|_F = \sqrt{\sum_{{i_1} =1}^{I_1} \sum_{{i_2} =1}^{I_2}\dots \sum_{{i_N} =1}^{I_N} \emph{x}_{i_1\times i_2\times\dots \times i_N}^2}$, and the $\ell_1$-norm as $\|{\cal X}\|_1 = \sum_{{i_1} =1}^{I_1} \sum_{{i_2} =1}^{I_2}\dots \sum_{{i_N} =1}^{I_N} \vert \emph{x}_{i_1\times i_2\times\dots \times i_N} \vert$. The $\nabla_{(i)}{\cal X}$ denotes the first-order derivative of the tensor $\cal X$ along the ${i}_{th}$ dimension.

\subsection{LR Tensor Recovery Modeling for Seismic Data Noise Suppression}
A seismic signal $\mathcal{X}  \in \mathbb{R}^{n_1\times n_2\times n_3}$ is usually contaminated by acquisition footprint and additive Gaussian noise, which is mathematically represented by
\begin{equation}
\cal Y=X+F+N,
\label{eq:1}
\end{equation} 
where $\mathcal{X}$ denotes the clean seismic data, while $\mathcal{Y}  \in \mathbb{R}^{n_1\times n_2\times n_3}$ represents the noisy seismic data. $\mathcal{F}  \in \mathbb{R}^{n_1\times n_2\times n_3}$ represents the footprint noise, which is usually characterized by linear stripes or grid patterns \cite{liu2021dictionary,qian2023improved}, while $\mathcal{N}  \in \mathbb{R}^{n_1\times n_2\times n_3}$ represents zero-mean additive Gaussian noise. The goal of seismic data noise suppression is to recover the clean seismic data from corrupted observations, which poses a highly ill-posed inverse problem due to the irreversible nature of the degradation process \cite{wang2015improved,li2022simultaneous,meng2024stochastic}. To address this challenge, priors capturing the intrinsic properties of seismic data are often employed to constrain the solution space, which can be formulated as follows:
\begin{equation}
\min_{\cal X, \cal F} \frac{1}{2}\|{\cal X} + {\cal F} - {\cal Y} \|_F^2 + \tau P({\cal X}) + \lambda P({\cal F}),
\label{eq:2}
\end{equation} 
where $\frac{1}{2}\|{\cal X} + {\cal F} - {\cal Y} \|_F^2$ denotes the data fidelity term, which provides the consistency between the recovered data and noisy observations. $\tau P({\cal X})$ represents the prior information of the clean seismic tensors, while $\lambda P({\cal F})$ serves as a regularizer based on the prior knowledge of the footprint noise. Here, $\tau$ and $\lambda$ are regularization parameters that balance the fidelity term and the sparsity regularization terms. Designing effective regularization terms is essential for seismic data recovery \cite{wang2020structure,liu2021high}. Recent studies have shown that LR tensor recovery models exhibit significant potential in suppressing seismic data noise \cite{feng2021low,feng2021seismic,qian2023improved,liu2024simultaneous}, with the UTV-LRTA method \cite{qian2023improved} emerging as one of the most promising approaches. This method combines a TNN with UTV priors, which is mathematically represented as,
\begin{equation}
\begin{aligned}
\hat{\cal X} = &\argmin_{\cal X}\frac{1}{2}\|{\cal X}  - {\cal Y}\|_F^2 + \tau \|{\cal X}\|_{*}\\
&+\lambda_1 \|\nabla_{(2)} {\cal X}\|_{1} + \lambda_2\|\nabla_{(1)}({\cal X}-{\cal Y})\|_{1},
\end{aligned}
\label{eq:5}
\end{equation}
where $\|\cdot\|_{*} $ denotes the TNN of seismic tensors \cite{lu2016tensor}, serving as a tensor sparsity measure to encode the LR structure underlying multidimensional seismic data. The term  $\lambda_1 \|\nabla_{(2)} {\cal X}\|_{1} + \lambda_2\|\nabla_{(1)}({\cal X}-{\cal Y})\|_{1}$ denotes the UTV regularizer, which captures the directional characteristic of acquisition footprint. Specially, the first term in the UTV regularizer promotes the local smoothness of $\cal X$, while the second term encourages the sparsity of the gradient of the acquisition footprint $\cal X-Y$ along its direction. This effectively removes the footprint noise and enhances the recovery of clean seismic data.

\subsection{Laplacian Scale Mixture (LSM) Modeling for Signal Recovery}
A sparse signal \(\textbf{\emph{x}} \in \mathbb{R}^{n}\) corrupted with Gaussian noise is usually recovered from the observed signal \(\textbf{\emph{y}}\) by solving the following $\ell_1$-norm minimization problem \cite{zha2020image,huang2017mixed},
\begin{equation}
\hat{\textbf{\emph{x}}}= \argmin\limits_{\textbf{\emph{x}}}\left\|{\textbf{\emph{y}}}-\textbf{\emph{x}}\right\|_2^2+\lambda\left\|\textbf{\emph{x}}\right\|_1.
\label{eq:6}
\end{equation}%
Solving Eq.~\eqref{eq:6} corresponds to performing maximum a posterior (MAP) inference for the signal $\textbf{\emph{x}}$ \cite{dong2015image}, assuming an independent and identically distributed (i.i.d.) Laplacian prior $P({\emph{x}}_i) = \frac{1}{2\theta_i}e^{-\frac{|{\emph{x}}_i|}{\theta_i}}$, where $\theta_i$ represents the standard deviation of ${\emph{x}}_i$. It can be easily verified that the regularization parameter $\lambda$ can be readily determined as $\lambda_i = \frac{2\boldsymbol{\sigma}_n^2}{\theta_i}$ \cite{zha2020benchmark}, where $\boldsymbol{\sigma}_n^2$ represents the variance of the approximated errors. However, the variance $\theta_i$ of each ${\emph{x}}_i$ is unknown and may be challenging to accurately estimate from the observation ${\textbf{\emph{y}}}$ because ${\textbf{\emph{y}}}$ is typically non-stationary and contaminated by noise. To address this issue, the LSM model \cite{dong2015low,xue2022laplacian} decomposes the signal $\textbf{\emph{x}}$ into a pointwise product of a Laplacian vector $\boldsymbol\beta$ and a positive hidden scalar multiplier $\boldsymbol\theta$, expressed as ${\emph{x}}_i = \theta_i\beta_i, \forall i$. Conditioned on $\theta_i$, ${\emph{x}}_i$ follows a Laplacian distribution with a standard deviation of $\theta_i$. Assuming that $\theta_i$ are i.i.d. and independent of $\beta_i$, the LSM prior for each ${\emph{x}}_i$ is formulated as,
\begin{equation}
P({\emph{x}}_i) = \int_{0}^\infty P({\emph{x}}_i|\theta_i)P(\theta_i)d\theta_i.
\label{eq:7}
\end{equation}%
Given an observation ${\textbf{\emph{y}}} = {\textbf{\emph{x}}} + \textbf{\emph{n}}$, where $\textbf{\emph{n}}$ represents additive Gaussian noise following the distribution $\textbf{\emph{n}} \sim \mathcal{N}(0, \boldsymbol{\sigma}_n^2)$, the sparse signal ${\textbf{\emph{x}}} $ and its hidden scalar parameter $\boldsymbol\theta$ can be jointly estimated using the following MAP estimator,
\begin{equation}
\begin{aligned}
(\hat{{\textbf{\emph{x}}}}, \hat{\boldsymbol\theta})& =\arg\max {\rm log}P({\textbf{\emph{y}}}|{\textbf{\emph{x}}}, \boldsymbol\theta) + {\rm log}P({\textbf{\emph{x}}}, \boldsymbol\theta)\\
&=\arg\max {\rm log}P({\textbf{\emph{y}}}|{\textbf{\emph{x}}}) + {\rm log}P({\textbf{\emph{x}}}|\boldsymbol\theta) + {\rm log}P(\boldsymbol\theta),
\label{eq:8}
\end{aligned}
\end{equation}
where $P({\textbf{\emph{y}}}|{\textbf{\emph{x}}})$ denotes the likelihood term that is modeled as a Gaussian distribution with a variance of $\boldsymbol{\sigma}_n^2$. The prior term $P({\textbf{\emph{x}}}|\boldsymbol\theta)$ is expressed as:
\begin{equation}
P({\textbf{\emph{x}}}|\boldsymbol\theta)= \prod\limits_{i} P({{\emph{x}}}_i|\theta_i) = \prod\limits_{i}\frac{1}{2\theta_i}{\rm exp} \left(-\frac{|{{\emph{x}}}_i|}{\theta_i}\right).
\label{eq:9}
\end{equation}%
By utilizing the LSM model, the concept of sparsity can be extended from the statistical modeling of the signal $\textbf{{\emph{x}}}$ to the definition of the sparse prior $P(\boldsymbol\theta)$. The prior is typically modeled as an i.i.d. noninformative Jeffrey's prior, $P(\theta_i )= \frac{1}{\theta_i}$ \cite{box2011bayesian}. Under this prior, Eq.~\eqref{eq:8} can be reformulated as,
\begin{equation}
(\hat{\textbf{\emph{x}}}, \hat{\boldsymbol\theta}) =\argmin\limits_{\textbf{\emph{x}}, \boldsymbol\theta}\|\textbf{\emph{y}} -\textbf{\emph{x}} \|_2^2 + 4\boldsymbol{\sigma}_n^2\sum\limits_i {\rm log} (\theta_i + \varepsilon)+ 2\boldsymbol{\sigma}_n^2\sum\limits_i \frac{|{{\emph{x}}}_i|}{\theta_i},
\label{eq:10}
\end{equation}
where $\varepsilon$ is a small positive constant. By using the LSM model, Eq.~\eqref{eq:10} can be rewritten as,
\begin{equation}
(\hat{\boldsymbol\beta}, \hat{\boldsymbol\theta}) =\argmin\limits_{\boldsymbol\beta, \boldsymbol\theta}\|\textbf{\emph{y}} -\boldsymbol\Lambda \boldsymbol\beta\|_2^2 + 4\boldsymbol{\sigma}_n^2\sum\limits_i {\rm log} (\theta_i + \varepsilon)+ 2\boldsymbol{\sigma}_n^2\sum\limits_i|\beta_i|,
\label{eq:11}
\end{equation}
where ${\textbf{\emph{x}}} = \boldsymbol\Lambda\boldsymbol\beta$, and  $\boldsymbol\Lambda = {\rm diag}(\theta_i)$ is a diagonal matrix representing the variance field. After solving Eq.~\eqref{eq:11} to obtain $\boldsymbol\beta$ and $\boldsymbol\theta$, the signal $\textbf{\emph{x}}$ can be reconstructed as $\hat{\textbf{\emph{x}}} =  \hat{\boldsymbol\Lambda} \hat{\boldsymbol\beta}$.

\section{Methodology}
\label{sec:4}

As mentioned above, most existing sparsity-based tensor recovery methods \cite{feng2021low,feng2021seismic,qian2023improved,liu2024simultaneous} for seismic data noise suppression enforce sparsity constraints on all regularization terms using soft-thresholding or hard-thresholding operators. However, these methods overlook the fact that real seismic data are non-stationary, making the variances of tensor coefficients unknown and difficult to accurately estimate from degraded seismic data, which leads to suboptimal noise suppression performance, as exemplified by Fig.~\ref{fig:0}(d). 

\subsection{TLSM Model for Seismic Data Noise Suppression}
In this paper, we propose an innovative TLSM approach for seismic data noise suppression, which integrates the LSM prior \cite{garrigues2010group} to enforce sparsity across all regularization terms, which can be formulated as,
\begin{equation}
\begin{aligned}
\hat{\cal X} = &\argmin_{\cal X}\frac{1}{2}\|{\cal X}  - {\cal Y}\|_F^2 + \tau \|{\cal X}\|_{\rm LSM,*}\\
&+\lambda_1 \|\nabla_{(2)} {\cal X}\|_{\rm LSM} + \lambda_2\|\nabla_{(1)}({\cal X}-{\cal Y})\|_{\rm LSM},
\end{aligned}
\label{eq:12}
\end{equation}
where $\|\cdot\|_{\rm LSM, *} $ and $\|\cdot\|_{\rm LSM} $ represent the LR and sparsity penalties, respectively, imposed using the LSM prior, as defined in Eq.~\eqref{eq:11}.  It is evident that, compared to the traditional LR tensor recovery model in Eq.~\eqref{eq:5}, the proposed TLSM approach jointly estimates both the variances and values of sparse tensor coefficients, achieving better denoising performance than many previous methods (see Section~\ref{sec:5} for more details). To highlight the advantages of our TLSM over existing methods, a comparative analysis is presented in Table~\ref{Tab:0}.
\begin{table}[!t]
\centering
\caption{A comparative analysis of the advantages and limitations between the proposed TLSM and existing state-of-the-art methods.}
\resizebox{0.49\textwidth}{!}				
{
\Huge
\begin{tabular}{|c|c|c|c|c|c|}
\hline
Methods        & \begin{tabular}[c]{@{}c@{}}LR \\ Regularization\end{tabular} & \begin{tabular}[c]{@{}c@{}}UTV \\ Regularization\end{tabular} & \begin{tabular}[c]{@{}c@{}}Adaptive \\ Sparse Prior\end{tabular} & \begin{tabular}[c]{@{}c@{}}Non-stationary \\ Noise\end{tabular} \\ \hline
DRR \cite{chen2016simultaneous}                         & $\checkmark$               & $\times$                             & $\times$                    & $\times$                   \\ \hline
SGRDL \cite{chen2021statistics}                      & $\times$                & $\times$                             & $\checkmark$                   & $\times$                   \\ \hline
FR-Net \cite{qian2023unsupervised}                     & $\times$                & $\checkmark$                            & $\times$                    & $\times$                   \\ \hline
DNLR \cite{xu2023deep}                       & $\checkmark$               & $\times$                             & $\times$                    & $\times$                   \\ \hline
S2S-WTV \cite{xu2023s2s}                    & $\checkmark$               & $\times$                             & $\times$                    & $\times$                   \\ \hline
UTV-LRTA \cite{qian2023improved}                   & $\checkmark$               & $\checkmark$                             & $\times$                    & $\times$                   \\ \hline
\textbf{TLSM}               & $\checkmark$               & $\checkmark$                            & $\checkmark$                   & $\checkmark$                  \\ \hline
\end{tabular}
}
\label{Tab:0}
\vspace{-2mm}
\end{table}

\subsection{Algorithm}
In this subsection, we develop an ADMM algorithm \cite{boyd2011distributed} to solve the proposed TLSM-based seismic data noise suppression problem.   Specifically, we introduce three auxiliary variables ${\cal Z} = {\cal X}$, ${\cal D}_{(2)} = \nabla_{(2)} {\cal X}$, and ${\cal D}_{(1)} = \nabla_{(1)}({\cal X}-{\cal Y})$. With these substitutions, Eq.~\eqref{eq:12} can be reformulated as the following constrained optimization problem,
\begin{equation}
\begin{aligned}
&\{\hat{\cal X}, \hat{\cal Z}, \hat{\cal D}_{(1)}, \hat{\cal D}_{(2)}\}= \argmin_{{\cal X}, {\cal Z}, {\cal D}_{(1)}, {\cal D}_{(2)}}\frac{1}{2}\|{\cal X} - {\cal Y}\|_F^2 + \tau \|{\cal Z}\|_{\rm LSM,*} \\
& \qquad \qquad \qquad \qquad \qquad + \lambda_1 \|{\cal D}_{(2)}\|_{\rm LSM} + \lambda_2 \|{\cal D}_{(1)}\|_{\rm LSM},\\
& { s. t.} \quad {\cal Z} = {\cal X},\  {\cal D}_{(2)} = \nabla_{(2)} {\cal X},\  {\cal D}_{(1)} = \nabla_{(1)}({\cal X}-{\cal Y}).
\end{aligned}
\label{eq:13}
\end{equation}
By invoking the ADMM algorithm, the optimization problem in Eq.~\eqref{eq:13} is transformed into seven iterative steps:
\begin{equation}
\begin{aligned}
&\hat{\cal X} \leftarrow \argmin_{{\cal X}} \frac{1}{2}\|{\cal X} - {\cal Y}\|_F^2 + \frac{a}{2}\|{\cal Z} - {\cal X}- {\cal B}\|_F^2\\
&\qquad \qquad + \frac{b}{2}\|{\cal D}_{(2)} - \nabla_{(2)} {\cal X}- {\cal B}_{(2)}\|_F^2\\
&\qquad \qquad + \frac{c}{2}\|{\cal D}_{(1)} - \nabla_{(1)}({\cal X}-{\cal Y})- {\cal B}_{(1)}\|_F^2,
\end{aligned}
\label{eq:131}
\end{equation}
\begin{equation}
\hat{\cal Z} \leftarrow \argmin_{{\cal Z}}  \frac{a}{2}\|{\cal Z} - {\cal X}- {\cal B}\|_F^2 + \tau \|{\cal Z}\|_{\rm LSM,*},
\label{eq:132}
\end{equation}
\begin{equation}
\begin{aligned}
&\hat{\cal D}_{(1)} \leftarrow \argmin_{{\cal D}_{(1)}} \frac{c}{2}\|{\cal D}_{(1)} - \nabla_{(1)}({\cal X}-{\cal Y})- {\cal B}_{(1)}\|_F^2\\
&\qquad \qquad + \lambda_1 \|{\cal D}_{(1)}\|_{\rm LSM},
\end{aligned}
\label{eq:133}
\end{equation}
\begin{equation}
\begin{aligned}
&\hat{\cal D}_{(2)} \leftarrow \argmin_{{\cal D}_{(2)}} \frac{b}{2}\|{\cal D}_{(2)} - \nabla_{(2)} {\cal X}- {\cal B}_{(2)}\|_F^2\\
&\qquad \qquad + \lambda_2 \|{\cal D}_{(2)}\|_{\rm LSM},
\end{aligned}
\label{eq:134}
\end{equation}
\begin{equation}
\hat{\cal B} \leftarrow {\cal B} - ({\cal Z} -{\cal X}),
\label{eq:135}
\end{equation}
\begin{equation}
\hat{\cal B}_{(1)} \leftarrow {\cal B}_{(1)} - ({\cal D}_{(1)} -\nabla_{(1)}({\cal X}-{\cal Y})),
\label{eq:136}
\end{equation}
\begin{equation}
\hat{\cal B}_{(2)} \leftarrow {\cal B}_{(2)} - ({\cal D}_{(2)}-\nabla_{(2)} {\cal X}),
\label{eq:137}
\end{equation}
where $\cal B$, $\cal {B}_{\text {(1)}}$, and $\cal {B}_{\text {(2)}}$ represent the Lagrangian multipliers, while  $a$, $b$, and $c$ are the balancing factors. The optimization of Eq.~\eqref{eq:13} consists of four subproblems, each of which can be solved independently with a closed-form solution, as detailed in the following subsections.

\subsubsection{${\cal X}$ Subproblem}
The ${\cal X}$ subproblem in Eq.~\eqref{eq:131} can be solved by independently addressing each slice ${\cal X}^{(k)}$ of the tensor $\cal X$, where $1\leq k \leq n_3$, as shown below,
\begin{equation}
\begin{aligned}
\hat{\cal X}^{(k)} = &\argmin_{{\cal X}^{(k)}} \frac{1}{2}\|{\cal X}^{(k)}  - {\cal Y}^{(k)}\|_F^2 + \frac{a}{2}\|{\cal Z}^{(k)} - {\cal X}^{(k)}- {\cal B}^{(k)}\|_F^2\\
&{}+ \frac{b}{2}\|{\cal D}_{(2)}^{(k)} - \nabla_{(2)} {\cal X}^{(k)}- {\cal B}_{(2)}^{(k)}\|_F^2\\
&{}+ \frac{c}{2}\|{\cal D}_{(1)}^{(k)} - \nabla_{(1)}({\cal X}^{(k)}-{\cal Y}^{(k)})- {\cal B}_{(1)}^{(k)}\|_F^2.
\end{aligned}
\label{eq:215}
\end{equation}
Since the objective function of Eq.~\eqref{eq:215} is differentiable, the optimality conditions for $\hat{\cal X}^{(k)}$ can be easily derived. By differentiating with respect to ${\cal X}^{(k)}$ and setting the result equal to zero, we obtain the following closed-form solution,
\begin{equation}
\begin{split}
\hat{\cal X}^{(k)} = &\left( a + b \nabla_{(2)}^T \nabla_{(2)} + c \nabla_{(1)}^T \nabla_{(1)} + {\cal I}^{(k)} \right)^{-1} \\
&\times\left(
\begin{aligned}
& a({\cal Z}^{(k)} - {\cal B}^{(k)}) + b \nabla_{(2)}^T ({\cal D}_{(2)}^{(k)} - {\cal B}_{(2)}^{(k)}) \\
& + c \nabla_{(1)}^T \left( {\cal D}_{(1)}^{(k)} + \nabla_{(1)} {\cal Y}^{(k)} - {\cal B}_{(1)}^{(k)} \right) + {\cal Y}^{(k)}
\end{aligned}
\right),
\end{split}
\label{eq:22}
\end{equation}
where ${\cal I}^{(k)}$ denotes the identity matrix. It is important to note that  Eq.~\eqref{eq:22} can be computed efficiently by leveraging the fast Fourier transform (FFT) algorithm \cite{chang2015anisotropic,chan2013constrained}.

\subsubsection{${\cal Z}$ Subproblem}
With the other variables held constant, the ${\cal Z}$  subproblem can be rewritten as,
\begin{equation}
\hat{\cal Z} = \argmin_{{\cal Z}}  \frac{a}{2}\|{\cal L} - {\cal Z}\|_F^2 + \tau \|{\cal Z}\|_{\rm LSM,*},
\label{eq:23}
\end{equation}
where ${\cal L} = {\cal X} +{\cal B}$. In this paper, we adopt a high effectively tensor singular value decomposition (t-SVD) \cite{zhang2014novel} method to tensor decomposition, and therefore, the ${\cal Z}$  subproblem in Eq.~\eqref{eq:23} can be transformed into solving the following problem,
\begin{equation}
\hat{\cal S} = \argmin_{{\cal S}}  \frac{a}{2}\|{\cal G} - {\cal S}\|_F^2 + \tau \|{\cal S}\|_{\rm LSM,*},
\label{eq:24}
\end{equation}
where ${\cal L} = {\cal U} \ast {\cal G} \ast {\cal V}^T$ and  ${\cal Z} = {\cal U} \ast {\cal S} \ast {\cal V}^T$.   $\cal U$ and $\cal V$ are orthogonal tensors, respectively. The symbol $\ast$ denotes the $t$-product \cite{zhang2014novel}. This decomposition is obtained by performing matrix SVDs in the Fourier domain \cite{kilmer2011factorization}. For more details on the $t$-product and $t$-SVD, refer to \cite{kilmer2011factorization,zhang2014novel}. Unlike the traditional $t$-SVD approach \cite{zhang2014novel}, which employs the soft-thresholding operator \cite{donoho1995noising}, we incorporate the LSM regularizer for ${\cal S}$, which can significantly enhance the estimation accuracy of both the sparse tensor coefficients and the hidden scalar parameters, leading to superior recovery results. To simplify notation in the following derivation, we represent all tensors in vector form. Specifically, we define $\boldsymbol{g}$ and $\boldsymbol{s}$ as the one-dimensional representations of $\cal G$ and $\cal S$, respectively. By using the LSM model \cite{dong2015low}, ${\cal Z}$ subproblem in Eq.~\eqref{eq:23} can be reformulated as,
\begin{equation}
\begin{aligned}
&(\hat{\boldsymbol{\alpha}},\hat{\boldsymbol{\theta}}) = \argmin_{{\boldsymbol{\alpha}}, {\boldsymbol{\theta}}}  \frac{a}{2}\|\boldsymbol{g} - \boldsymbol\Lambda \boldsymbol{\alpha}\|_2^2 + \sqrt{2}\tau\sum_i|\alpha_i|\\
&\qquad \qquad +2\tau\sum_i{\rm log} (\theta_i + \varepsilon),
\end{aligned}
\label{eq:25}
\end{equation}
where $\boldsymbol{s} = \boldsymbol\Lambda \boldsymbol\alpha$, and  $\boldsymbol\Lambda = {\rm diag}(\theta_i)$.   It can be observed that the ${\cal Z}$ subproblem is reformulated into solving the $\boldsymbol{\alpha}$ and $\boldsymbol{\theta}$ subproblems. The detailed derivation process for solving the $\boldsymbol{\alpha}$ and $\boldsymbol{\theta}$ subproblems is provided below.

\begin{center}
\begin{algorithm}[!t]
\caption{The Proposed TLSM Algorithm for Seismic Data Noise Suppression}
\begin{algorithmic}[1]
\REQUIRE Noisy seismic tensor ${\cal Y} \in \mathbb{R}^{n_1\times n_2\times n_3}$, parameters $a, b, c, \tau, \lambda_1, \lambda_2$, and maximum iteration number $T$.
\STATE \textbf{Initialization:} Initialize $\cal B$, $\cal {B}_{\text {(1)}}$, $\cal {B}_{\text {(2)}}$, $\cal {D}_{\text {(1)}}$, and $\cal {D}_{\text {(2)}}$ as zero tensors. Set the initial estimate as ${\cal Z} = {\cal Y}$.
\FOR{$t = 1, \dots, T$}
\FOR{$k = 1, \dots, n_3$}	
\STATE Compute $\hat{{\cal X}}^{(k)}$ using Eq.~\eqref{eq:22};
\ENDFOR
\STATE Compute $\hat{\boldsymbol{\theta}}$ using Eq.~\eqref{eq:28};
\STATE Compute $\hat{\boldsymbol{\alpha}}$ using Eq.~\eqref{eq:31};
\STATE Compute $\hat{{\cal Z}}$ using Eq.~\eqref{eq:315};
\STATE Compute $\hat{\boldsymbol{\beta}}$ similarly to $\hat{\boldsymbol{\alpha}}$;
\STATE Compute $\hat{\boldsymbol{o}}$ similarly to $\hat{\boldsymbol{\theta}}$;
\STATE Compute $\hat{{\cal D}}_{(1)}$ by folding the vector $\hat{\boldsymbol{d}} = \hat{\boldsymbol{\Phi}} \hat{\boldsymbol{\beta}}$;
\STATE Compute $\hat{{\cal D}}_{(2)}$ similarly to $\hat{{\cal D}}_{(1)}$;
\STATE Compute $\hat{{\cal B}}$ using Eq.~\eqref{eq:135};
\STATE Compute $\hat{{\cal B}}_{(1)}$ using Eq.~\eqref{eq:136};
\STATE Compute $\hat{{\cal B}}_{(2)}$ using Eq.~\eqref{eq:137};
\ENDFOR
\STATE \textbf{Output:} Reconstructed seismic data $\hat{\cal X}$.
\end{algorithmic}
\label{algo:1}
\end{algorithm}
\end{center}

\begin{table*}[!t]
\centering
\caption{Average noise suppression results of PSNR ($\textnormal{d}$B) and SSIM metrics by DRR \cite{chen2016simultaneous}, SGRDL \cite{chen2021statistics}, FR-Net \cite{qian2023unsupervised}, DNLR \cite{xu2023deep}, S2S-WTV \cite{xu2023s2s}, UTV-LRTA \cite{qian2023improved} and the proposed TLSM methods on the synthetic seismic dataset with varying noise levels.}
\centering
\tiny
\scriptsize
\centering
\resizebox{1\textwidth}{!}
{
\begin{tabular}{|cc|c|c|c|c|c|c|c|c|}
\hline
\multicolumn{2}{|c|}{Noise Levels}                                     & \multirow{2}{*}{Index} & \multirow{2}{*}{DDR \cite{chen2016simultaneous}} & \multirow{2}{*}{SGRDL \cite{chen2021statistics}} & \multirow{2}{*}{FR-Net \cite{qian2023unsupervised}} & \multirow{2}{*}{DNLR \cite{xu2023deep}} & \multirow{2}{*}{S2S-WTV \cite{xu2023s2s}} & \multirow{2}{*}{UTV-LRTA \cite{qian2023improved}} & \multirow{2}{*}{\textbf{TLSM}} \\ \cline{1-2}
\multicolumn{1}{|c|}{Footprint}              & Gaussian                &                        &                      &                        &                         &                       &                          &                           &                                \\ \hline
\multicolumn{1}{|c|}{\multirow{8}{*}{$F = 0.1$}} & \multirow{2}{*}{$\sigma$ = 0.01} & PSNR                   & 23.05                & 37.22                  & 41.00                   & 37.88                 & 40.75                    & 38.95                     & \textbf{44.06}                 \\ \cline{3-10}
\multicolumn{1}{|c|}{}                       &                         & SSIM                   & 0.9198               & 0.9270                 & 0.9686                  & 0.9722                & 0.9595                   & 0.9685                    & \textbf{0.9927}                \\ \cline{2-10}
\multicolumn{1}{|c|}{}                       & \multirow{2}{*}{$\sigma$ = 0.02} & PSNR                   & 23.05                & 37.05                  & 35.21                   & 37.81                 & 40.73                    & 35.11                     & \textbf{41.49}                 \\ \cline{3-10}
\multicolumn{1}{|c|}{}                       &                         & SSIM                   & 0.9198               & 0.9222                 & 0.8935                  & 0.9679                & 0.9560                   & 0.8957                    & \textbf{0.9801}                \\ \cline{2-10}
\multicolumn{1}{|c|}{}                       & \multirow{2}{*}{$\sigma$ = 0.03} & PSNR                   & 23.04                & 36.79                  & 31.83                   & 37.52                 & \textbf{40.29}           & 31.97                     & 38.92                          \\ \cline{3-10}
\multicolumn{1}{|c|}{}                       &                         & SSIM                   & 0.9198               & 0.9201                 & 0.8030                  & 0.9596                & 0.9532                   & 0.8078                    & \textbf{0.9606}                \\ \cline{2-10}
\multicolumn{1}{|c|}{}                       & \multirow{2}{*}{$\sigma$ = 0.04} & PSNR                   & 23.04                & 36.56                  & 29.37                   & 35.68                 & \textbf{40.04}           & 29.62                     & 36.80                          \\ \cline{3-10}
\multicolumn{1}{|c|}{}                       &                         & SSIM                   & 0.9198               & 0.9199                 & 0.7162                  & \textbf{0.9548}       & 0.9519                   & 0.8826                    & 0.9360                         \\ \hline
\multicolumn{1}{|c|}{\multirow{8}{*}{$F = 0.2$}} & \multirow{2}{*}{$\sigma$ = 0.01} & PSNR                   & 21.97                & 34.79                  & 39.97                   & 30.53                 & 34.53                    & 38.81                     & \textbf{43.80}                 \\ \cline{3-10}
\multicolumn{1}{|c|}{}                       &                         & SSIM                   & 0.7719               & 0.8972                 & 0.9676                  & 0.7971                & 0.8349                   & 0.9685                    & \textbf{0.9918}                \\ \cline{2-10}
\multicolumn{1}{|c|}{}                       & \multirow{2}{*}{$\sigma$ = 0.02} & PSNR                   & 21.97                & 34.50                  & 35.06                   & 30.12                 & 34.25                    & 34.97                     & \textbf{41.36}                 \\ \cline{3-10}
\multicolumn{1}{|c|}{}                       &                         & SSIM                   & 0.7719               & 0.8928                 & 0.8916                  & 0.7942                & 0.8321                   & 0.8954                    & \textbf{0.9800}                \\ \cline{2-10}
\multicolumn{1}{|c|}{}                       & \multirow{2}{*}{$\sigma$ = 0.03} & PSNR                   & 21.97                & 32.98                  & 31.48                   & 30.04                 & 33.96                    & 31.90                     & \textbf{38.70}                 \\ \cline{3-10}
\multicolumn{1}{|c|}{}                       &                         & SSIM                   & 0.7719               & 0.8756                 & 0.7994                  & 0.7900                & 0.8270                   & 0.8072                    & \textbf{0.9598}                \\ \cline{2-10}
\multicolumn{1}{|c|}{}                       & \multirow{2}{*}{$\sigma$ = 0.04} & PSNR                   & 21.96                & 32.87                  & 29.34                   & 28.72                 & 33.86                    & 29.62                     & \textbf{36.63}                 \\ \cline{3-10}
\multicolumn{1}{|c|}{}                       &                         & SSIM                   & 0.7718               & 0.8747                 & 0.7122                  & 0.7677                & 0.8253                   & 0.7220                    & \textbf{0.9353}                \\ \hline
\multicolumn{1}{|c|}{\multirow{8}{*}{$F = 0.5$}} & \multirow{2}{*}{$\sigma$ = 0.01} & PSNR                   & 20.31                & 27.05                  & 30.57                   & 19.64                 & 24.04                    & 38.70                     & \textbf{41.59}                 \\ \cline{3-10}
\multicolumn{1}{|c|}{}                       &                         & SSIM                   & 0.5145               & 0.7768                 & 0.9043                  & 0.4208                & 0.5422                   & 0.9684                    & \textbf{0.9853}                \\ \cline{2-10}
\multicolumn{1}{|c|}{}                       & \multirow{2}{*}{$\sigma$ = 0.02} & PSNR                   & 20.31                & 26.44                  & 29.51                   & 19.64                 & 23.56                    & 34.87                     & \textbf{39.90}                 \\ \cline{3-10}
\multicolumn{1}{|c|}{}                       &                         & SSIM                   & 0.5145               & 0.7763                 & 0.8302                  & 0.4195                & 0.5363                   & 0.8954                    & \textbf{0.9739}                \\ \cline{2-10}
\multicolumn{1}{|c|}{}                       & \multirow{2}{*}{$\sigma$ = 0.03} & PSNR                   & 20.31                & 26.42                  & 28.11                   & 19.62                 & 23.45                    & 31.83                     & \textbf{38.04}                 \\ \cline{3-10}
\multicolumn{1}{|c|}{}                       &                         & SSIM                   & 0.5145               & 0.7730                 & 0.7460                  & 0.4149                & 0.5330                   & 0.8072                    & \textbf{0.9563}                \\ \cline{2-10}
\multicolumn{1}{|c|}{}                       & \multirow{2}{*}{$\sigma$ = 0.04} & PSNR                   & 20.31                & 25.60                  & 26.33                   & 19.31                 & 23.22                    & 29.58                     & \textbf{36.19}                 \\ \cline{3-10}
\multicolumn{1}{|c|}{}                       &                         & SSIM                   & 0.5145               & 0.7656                 & 0.6676                  & 0.4047                & 0.5304                   & 0.7214                    & \textbf{0.9338}                \\ \hline
\multicolumn{2}{|c|}{\multirow{2}{*}{Average}}                         & PSNR                   & 21.77                & 32.36                  & 32.31                   & 28.88                 & 32.72                    & 33.83                     & \textbf{39.79}                 \\ \cline{3-10}
\multicolumn{2}{|c|}{}                                                 & SSIM                   & 0.7354               & 0.8601                 & 0.8250                  & 0.7220                & 0.7735                   & 0.8482                    & \textbf{0.9655}                \\ \hline
\end{tabular}
}
\label{Tab:1}
\vspace{-2mm}
\end{table*}

\paragraph{$\boldsymbol{\theta}$ Subproblem}
Given an initial estimate of $\boldsymbol{\alpha}$, the $\boldsymbol{\theta}$ subproblem in Eq.~\eqref{eq:25} simplifies to,
\begin{equation}
\hat{\boldsymbol{\theta}} = \argmin_{\boldsymbol{\theta}} \frac{a}{2}\|\boldsymbol{g} - \boldsymbol\Lambda \boldsymbol{\alpha}\|_2^2 + 2\tau\sum_i{\rm log} (\theta_i + \varepsilon).
\label{eq:26}
\end{equation}
It can be easily proven that Eq.~\eqref{eq:26} can be decomposed into a sequence of scalar minimization problems,
\begin{equation}
\hat{\theta}_i = \argmin\limits_{\theta_i} r_i \theta_i^2 + p_i\theta_i + 2\tau {\rm log}(\theta_i  + \varepsilon),
\label{eq:27}
\end{equation}
where $r_i = \frac{a}{2}\alpha^2_i$, and $p_i = -a g_i \alpha_i$. Therefore, Eq.~\eqref{eq:27} can be solved by setting $\frac{df(\theta_i)}{d\theta_i}=0$, where $f(\theta_i)$ denotes the right-hand side of Eq.~\eqref{eq:27}. Consequently, the solution of Eq.~\eqref{eq:27} is given by
\begin{equation}
\hat{\theta}_i =\left\{
\begin{aligned}
&0, \ \ \  {\rm if} \ \ (p_i^2-16r_i\tau)/16r_i^2<0\\
& \varphi_i, \ \  {\rm otherwise}.
\end{aligned}
\right.
\label{eq:28}
\end{equation}
Here, $ \varphi_i = {\rm argmin}_{\theta_i}\{f(0), f(\theta_{i,1}), f(\theta_{i,2})\}$, where $\theta_{i,1}$ and $\theta_{i,2}$ represent two stationary points of $f(\theta_{i}$), namely,
\begin{align}
\theta_{i,1} = -\frac{p_i}{4r_i}+\sqrt{\frac{p_i^2-16r_i\tau}{16r_i^2}},\nonumber\\
\theta_{i,2} = -\frac{p_i}{4r_i}-\sqrt{\frac{p_i^2-16r_i\tau}{16r_i^2}}.
\label{eq:29}
\end{align}
\paragraph{$\boldsymbol{\alpha}$ Subproblem}
By keeping the variable $\boldsymbol{\theta}$ fixed, the $\boldsymbol{\alpha}$ subproblem in Eq.~\eqref{eq:25} simplifies to,
\begin{equation}
\hat{\boldsymbol{\alpha}} = \argmin_{\boldsymbol{\alpha}}  \frac{a}{2}\|\boldsymbol{g} - \boldsymbol\Lambda \boldsymbol{\alpha}\|_2^2 + \sqrt{2}\tau\sum_i|\alpha_i|,
\label{eq:30}
\end{equation}
which admits the following closed-form solution,
\begin{equation}
\hat{\alpha}_i = {\rm Soft}({\frac{g_i}{\theta_i}, \tau_i}),
\label{eq:31}
\end{equation}
where ${\rm Soft} (\cdot)$ denotes the standard soft-thresholding function \cite{donoho1995noising} with a threshold of $\tau_i = \sqrt{2}\tau/a$.  Once the solutions for $\boldsymbol{\alpha}$ and $\boldsymbol{\theta}$ are obtained, we can compute $\hat{\boldsymbol{s}} = \hat{\boldsymbol\Lambda} \hat{\boldsymbol\alpha}$, and thus $\hat{\cal S}$ can obtained by folding $\hat{\boldsymbol{s}}$. Finally, the solution for ${\cal Z}$ is obtained as,
\begin{equation}
\hat{\cal Z} = {\cal U} \ast \hat{\cal S} \ast {\cal V}^T.
\label{eq:315}
\end{equation}

\subsubsection{${\cal D}_{(1)}$ Subproblem}
For fixed values of other variables, the ${\cal D}_{(1)}$ subproblem in Eq.~\eqref{eq:133} can be rewritten as,
\begin{equation}
\hat{\cal D}_{(1)} = \argmin_{{\cal D}_{(1)}} \frac{c}{2}\|{\cal K}_{(1)} - {\cal D}_{(1)}\|_F^2 + \lambda_1 \|{\cal D}_{(1)}\|_{\rm LSM},
\label{eq:32}
\end{equation}
where ${\cal K}_{(1)} = \nabla_{(1)}({\cal X}-{\cal Y})+ {\cal B}_{(1)}$. For convenience in the following derivation, we represent ${\cal D}_{(1)}$ and ${\cal K}_{(1)}$ in vector form, denoted by $\boldsymbol{d}$ and $\boldsymbol{k}$, respectively. Hence, Eq.~\eqref{eq:32} can be expressed as

\begin{equation}
\hat{\boldsymbol{d}} = \argmin_{\boldsymbol{d}} \frac{c}{2}\|\boldsymbol{k} - \boldsymbol{d}\|_2^2 + \lambda_1 \|\boldsymbol{d}\|_{\rm LSM},
\label{eq:33}
\end{equation}
By invoking the LSM model \cite{dong2015low}, ${\cal D}_{(1)}$ subproblem can be  reformulated as,
\begin{equation}
\begin{aligned}
&(\hat{\boldsymbol\beta},\hat{\boldsymbol{o}}) = \argmin_{\boldsymbol\beta, \boldsymbol{o}} \frac{c}{2}\|\boldsymbol{k} - \boldsymbol\Phi \boldsymbol\beta\|_2^2 + \sqrt{2}\lambda_1\sum_i|\beta_i|\\
&\qquad \qquad +2\lambda_1\sum_i{\rm log} (o_i + \varepsilon),
\end{aligned}
\label{eq:34}
\end{equation}
where $\boldsymbol{d} = \boldsymbol\Phi \boldsymbol\beta$, and $\boldsymbol\Phi = {\rm diag}(o_i)$.   It can be seen that the same computational scheme can be used to solve the $\boldsymbol{\theta}$ subproblem via Eq.~\eqref{eq:29} for the $\boldsymbol\beta$ subproblem, while also applying the same scheme to solve the $\boldsymbol{\alpha}$ subproblem via Eq.~\eqref{eq:31} for the $\boldsymbol{o}$ subproblem. Once the solutions for $\boldsymbol\beta$ and $\boldsymbol{o}$ are obtained, we can compute $\hat{\boldsymbol{d}} = \hat{\boldsymbol\Phi} \hat{\boldsymbol\beta}$, and consequently, $\hat{{\cal D}}_{(1)}$ is obtained by folding $\hat{\boldsymbol{d}}$.

\subsubsection{${\cal D}_{(2)}$ Subproblem}
For fixed other variables, the ${\cal D}_{(2)}$  subproblem is described as,
\begin{equation}
\hat{\cal D}_{(2)} = \argmin_{{\cal D}_{(2)}} \frac{c}{2}\|{\cal H}_{(2)} - {\cal D}_{(2)}\|_F^2 + \lambda_2 \|{\cal D}_{(2)}\|_{\rm LSM},
\label{eq:41}
\end{equation}
where ${\cal H}_{(2)} = \nabla_{(2)} {\cal X} + {\cal B}_{(2)}$. The ${\cal D}_{(2)}$ subproblem can be solved in a manner similar to the solution of the ${\cal D}_{(1)}$ subproblem in Eq.~\eqref{eq:32}, with the details omitted for brevity.

Building upon the aforementioned procedures, an effective solution for each subproblem is attained by invoking the ADMM algorithm \cite{boyd2011distributed}, making the proposed TLSM seismic data noise suppression algorithm more stable and practical. The complete procedure of the proposed TLSM-based seismic data noise suppression is summarized in Algorithm \ref{algo:1}, providing a clear and structured framework for implementation.

\begin{figure*}[!t]
\centering
\vspace{-2mm}
\begin{minipage}[b]{1\linewidth}
{\includegraphics[width= 1\textwidth]{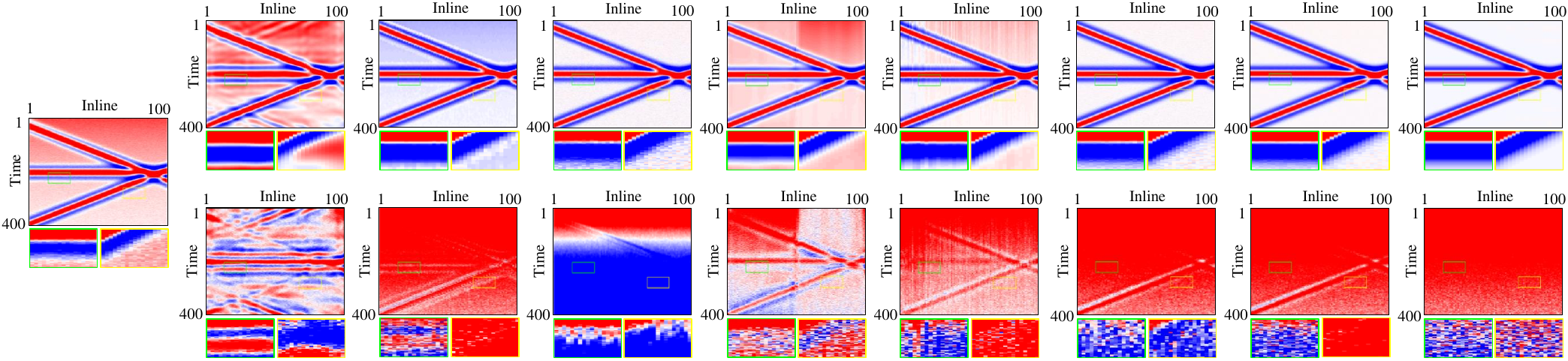}
\par\vspace{2mm}
\includegraphics[width= 1\textwidth]{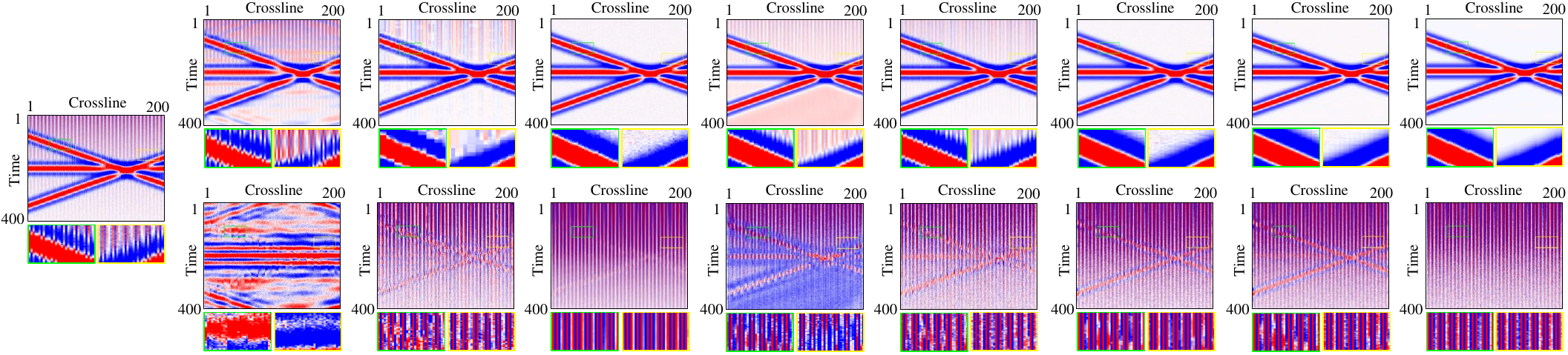}
\par\vspace{2mm}
\includegraphics[width= 1\textwidth]{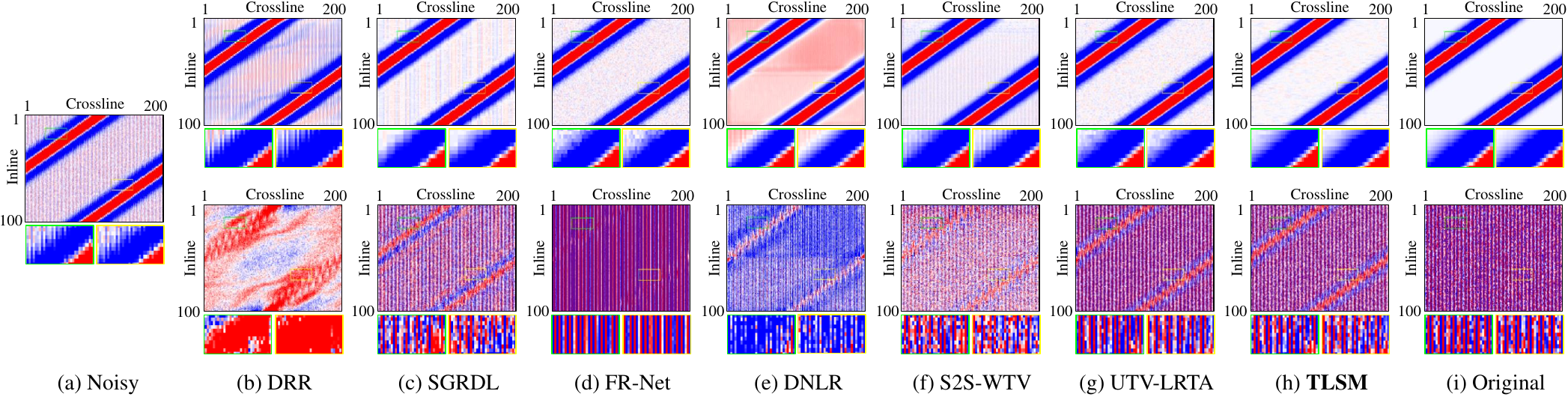}}
\end{minipage}
\vspace{-6mm}
\caption{Noise suppression results of seismic data (first, third, and fifth rows) and the corresponding removed noise (second, fourth, and sixth rows) using different methods on the synthetic data with size $100\times200\times400$. (a) Noisy seismic data ($F = 0.2, \sigma = 0.01$); (b) DRR \cite{chen2016simultaneous} (21.15dB); (c) SGRDL \cite{chen2021statistics} (35.41dB); (d) FR-Net \cite{qian2023unsupervised} (39.31dB); (e) DNLR \cite{xu2023deep} (27.79dB); (f) S2S-WTV \cite{xu2023s2s} (35.69dB); (g) UTV-LRTA \cite{qian2023improved} (39.58dB); (h) \textbf{TLSM} (\textbf{45.22dB}); (i) Original seismic data.}
\label{fig:1}
\vspace{-2mm}
\end{figure*}

\begin{table*}[!t]
\centering
\caption{Average running time (in seconds) of different methods for seismic data noise suppression.}
\centering
\tiny
\centering
\resizebox{1\textwidth}{!}
{
\begin{tabular}{|c|c|c|c|c|c|c|c|}
\hline
\textbf{Methods} & DRR\cite{chen2016simultaneous}     & SGRDL\cite{chen2021statistics}  & FR-Net\cite{qian2023unsupervised} & DNLR\cite{xu2023deep}    & S2S-WTV\cite{xu2023s2s} & UTV-LRTA\cite{qian2023improved} & \textbf{TLSM}   \\ \hline
\textbf{Time}    & 1036.52 & 94.70 & 110.00 & 1093.68 & 993.31 & 124.91   & \textbf{60.68} \\ \hline
\end{tabular}
}
\label{Tab:2}
\vspace{-2mm}
\end{table*}

\section{Experimental Results}
\label{sec:5}
In this section, we conduct extensive experiments to verify the effectiveness of the proposed TLSM model for seismic data noise suppression. The source code for the TLSM algorithm is available at: \url{https://github.com/pansirui/TLSM_Seismic_Denoising_Demo}.

\emph{Implementation Details:} We compare the proposed TLSM algorithm with six state-of-the-art seismic data noise suppression approaches, including three model-based approaches (DRR \cite{chen2016simultaneous}, SGRDL \cite{chen2021statistics} and UTV-LRTA \cite{qian2023improved}) and three deep-learning based approaches (FR-Net \cite{qian2023unsupervised}, DNLR \cite{xu2023deep} and S2S-WTV \cite{xu2023s2s}). All parameters are fine-tuned either by default settings or according to the guidelines provided in the respective papers to ensure optimal performance. The experiments are performed on a PC with an Inter(R) Core(TM) i7-12700K 3.60 GHz CPU and an NVIDIA GeForce RTX 4090 GPU. The MATLAB R2023a platform with CPU calculation is employed for implementing DRR, SGRDL, UTV-LRTA and our method. On the other hand, PyTorch 1.12.1 is utilized for implementing DNLR and S2S-WTV, and Tensorflow 2.5.0 is used for implementing FR-Net, with both CPU and GPU computation supported.

In the proposed TLSM algorithm, we empirically set the parameters $a, b, c, \tau, \lambda_1,$ and $\lambda_2$ to $(4, 0.2, 1, 0.5, 0.05, 1)$, $(1, 0.05, 1, 0.1, 10, 1)$, and $(0.1, 10, 1, 0.1, 10, 1)$ for the synthetic dataset, the Penobscot-3D dataset, and the Kerry-3D dataset, respetively. The number of iterations $T$ is set to 20. A more detailed discussion on the parameter settings of the proposed TLSM-based seismic data noise suppression algorithm can be found in subsection~\ref{sec:5.4.1}.

\emph{Performance Evaluation Metrics:} To comprehensively evaluate the noise suppression performance of all competing methods, two well-known quantitative metrics are employed: peak signal-to-noise ratio (PSNR) and structural similarity index (SSIM) \cite{wang2004image}. It is worth noting that higher PSNR and SSIM values correspond to better noise attenuation performance.

\begin{figure*}[!t]
\centering
\begin{minipage}[b]{1\linewidth}
{\includegraphics[width= 1\textwidth]{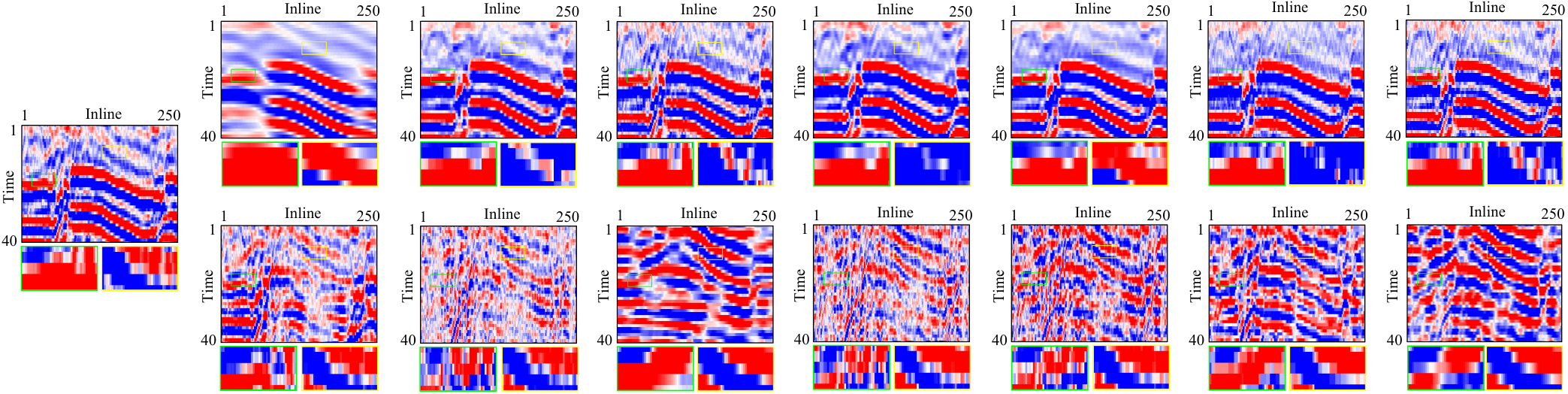}
\par\vspace{2mm}
\includegraphics[width= 1\textwidth]{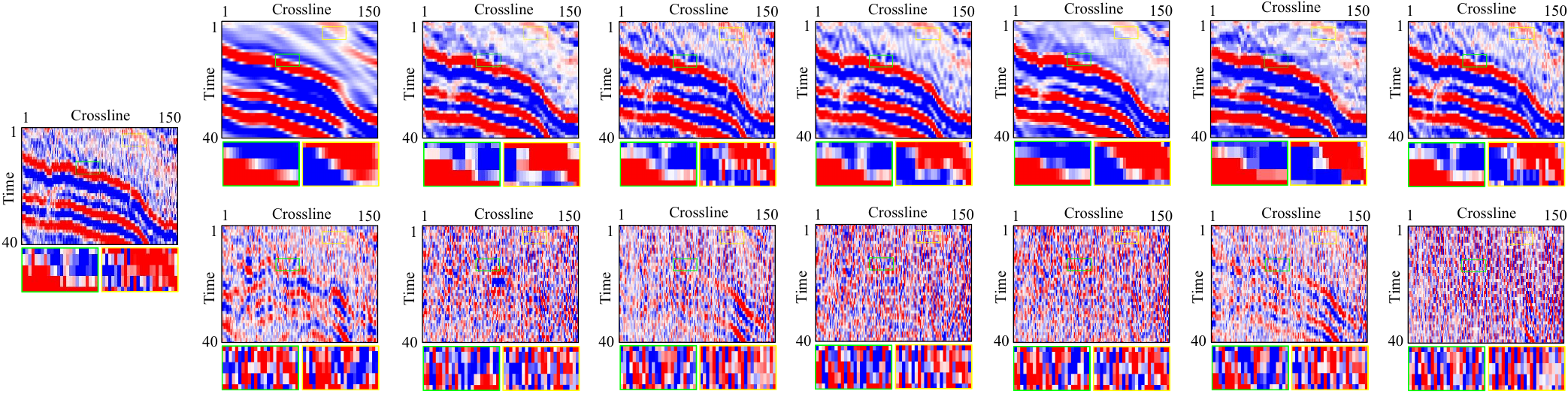}
\par\vspace{2mm}
\includegraphics[width= 1\textwidth]{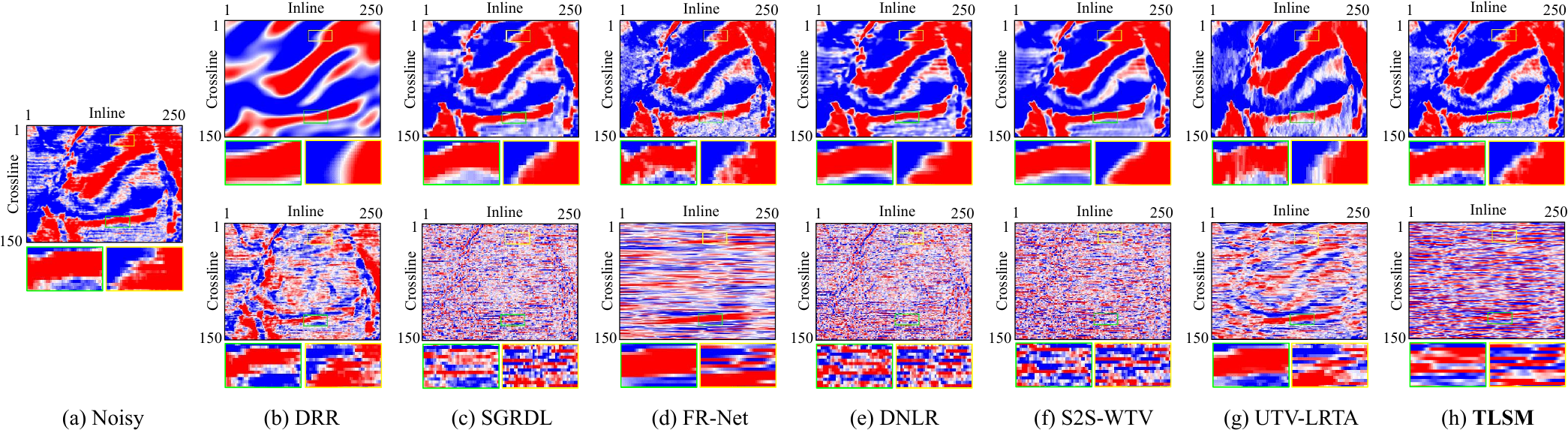}}
\end{minipage}
\vspace{-6mm}
\caption{Noise suppression results of seismic data (first, third, and fifth rows) and the corresponding removed noise (second, fourth, and sixth rows) using different methods on the Penobscot-3D dataset. (a) Noisy seismic data; (b) DRR \cite{chen2016simultaneous}; (c) SGRDL \cite{chen2021statistics}; (d) FR-Net \cite{qian2023unsupervised}; (e) DNLR \cite{xu2023deep}; (f) S2S-WTV \cite{xu2023s2s}; (g) UTV-LRTA \cite{qian2023improved}; (h) \textbf{TLSM}.}
\label{fig:2}
\vspace{-2mm}
\end{figure*}

\begin{figure*}[!t]
\centering
\begin{minipage}[b]{1\linewidth}
{\includegraphics[width= 1\textwidth]{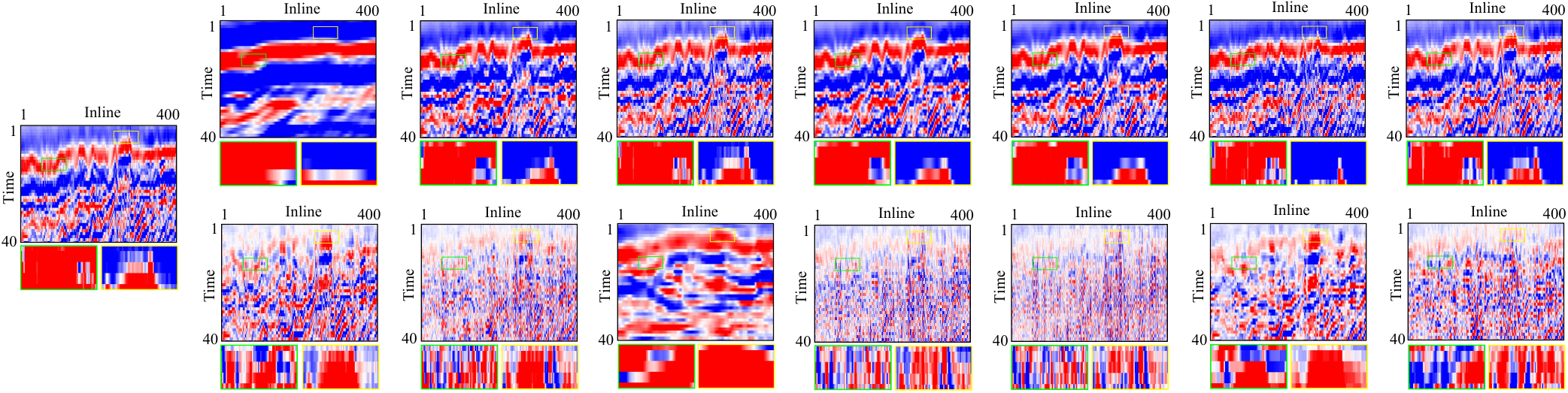}
\par\vspace{2mm}
\includegraphics[width= 1\textwidth]{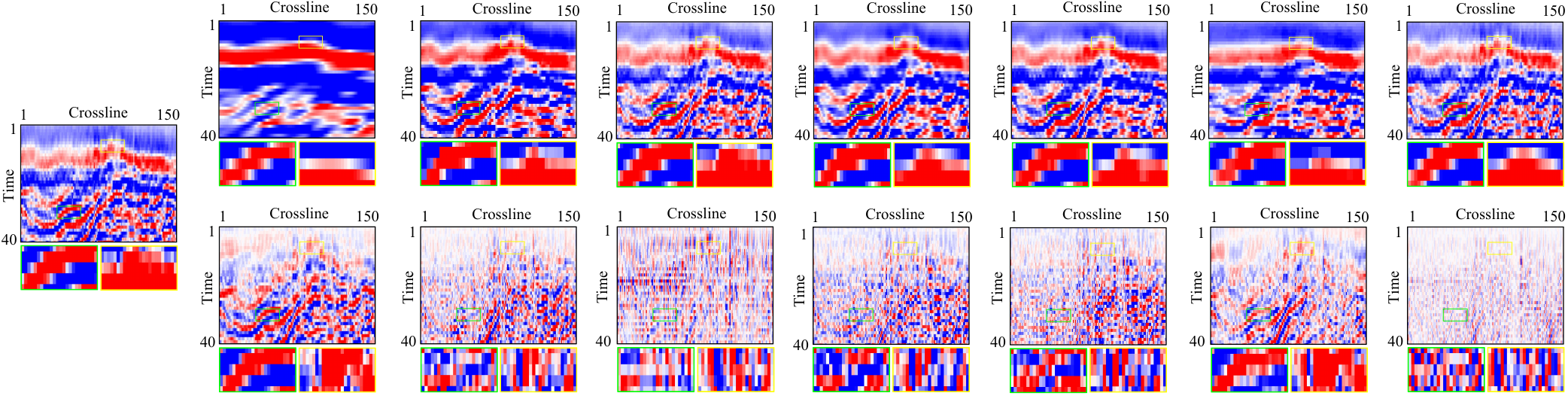}
\par\vspace{2mm}
\includegraphics[width= 1\textwidth]{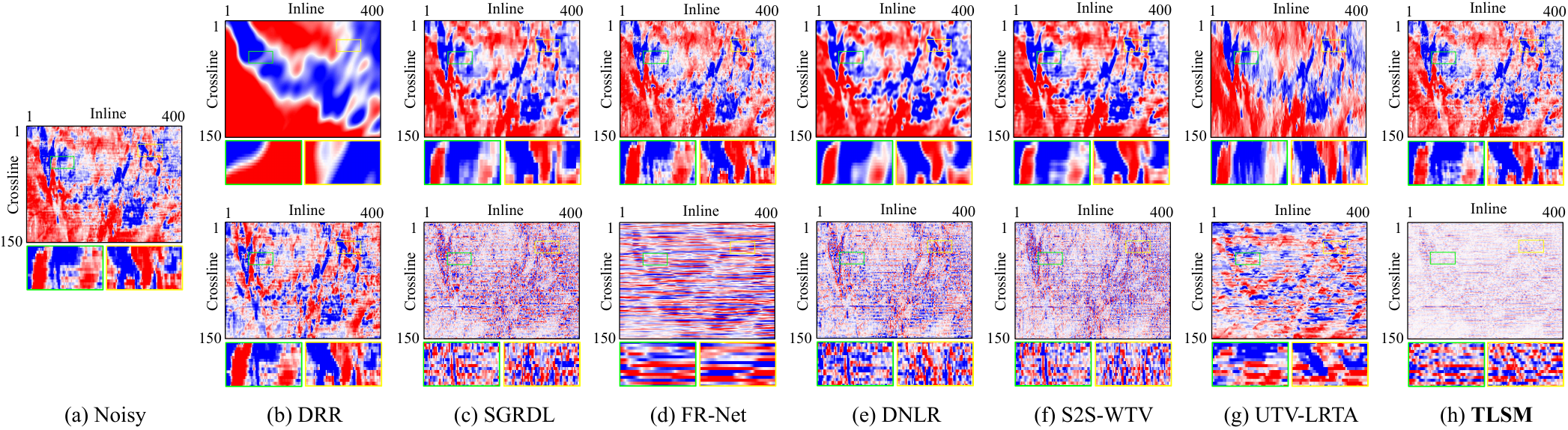}}
\end{minipage}
\vspace{-6mm}
\caption{Noise suppression results of seismic data (first, third, and fifth rows) and the corresponding removed noise (second, fourth, and sixth rows) using different methods on the Kerry-3D dataset. (a) Noisy seismic data; (b) DRR \cite{chen2016simultaneous}; (c) SGRDL \cite{chen2021statistics}; (d) FR-Net \cite{qian2023unsupervised}; (e) DNLR \cite{xu2023deep}; (f) S2S-WTV \cite{xu2023s2s}; (g) UTV-LRTA \cite{qian2023improved}; (h) \textbf{TLSM}.}
\label{fig:3}
\vspace{-2mm}
\end{figure*}

\subsection{Synthetic Data Experiments}

\emph{Testing Datasets:} 
The synthetic seismic data are generated using the code provided in \cite{chen2021statistics}, which includes three linear cross events. A sampling interval of 2 ms and a 10 Hz Ricker wavelet are chosen. The seismic data are then normalized to the range $[-1, +1]$. Four data dimension configurations are considered, with sizes of $40\times200\times400$, $100\times200\times400$, $200\times200\times400$, and $400\times200\times400$. The three dimensions of the seismic data represent inline, crossline, and time, respectively. To simulate noisy conditions, mixtures of the time-decaying footprint noise and the Gaussian noise are added to the synthetic seismic data. Different noise levels are created by varying the maximum amplitude $F$ of the footprint noise (with values 0.1, 0.2, and 0.5) and the standard deviation $\sigma$ of the Gaussian noise (with values 0.01, 0.02, 0.03, and 0.04), resulting in a total of 12 distinct noise configurations.

\begin{table*}[!t]
\centering
\caption{Average no-reference quality metrics on field seismic datasets.}
\centering
\tiny
\scriptsize
\centering
\resizebox{1\textwidth}{!}
{
\begin{tabular}{|c|c|c|c|c|c|c|c|c|}
\hline
Dataset                       & Index   & DRR \cite{chen2016simultaneous}   & SGRDL \cite{chen2021statistics} & FR-Net \cite{qian2023unsupervised}  & DNLR \cite{xu2023deep} & S2S-WTV \cite{xu2023s2s} & UTV-LRTA \cite{qian2023improved} & \textbf{TLSM}  \\ \hline
\multirow{3}{*}{Penobscot-3D} & BRISQUE \cite{mittal2012no} & 53.46 & 43.34 & 29.37  & 44.27 & 48.98   & 36.12    & \textbf{27.38} \\ \cline{2-9}
                              & NIQE \cite{mittal2012making}   & 9.32  & 10.92 & 7.58   & 9.80  & 6.55    & 9.73     & \textbf{5.66}  \\ \cline{2-9}
                              & PIQE \cite{venkatanath2015blind}   & 87.19 & 60.47 & 28.91  & 80.63 & 81.42   & 39.24    & \textbf{27.13} \\ \hline
\multirow{3}{*}{Kerry-3D}     & BRISQUE \cite{mittal2012no} & 61.78 & 43.46 & 36.79  & 51.52 & 44.96   & 35.95    & \textbf{35.83} \\ \cline{2-9}
                              & NIQE \cite{mittal2012making}   & 8.04  & 12.24 & 5.38   & 8.29  & 8.00    & 7.59     & \textbf{5.34}  \\ \cline{2-9}
                              & PIQE \cite{venkatanath2015blind}   & 91.40 & 62.30 & 19.92  & 85.58 & 77.49   & 38.87    & \textbf{18.23} \\ \hline
\end{tabular}
}
\label{Tab:25}
\vspace{-4mm}
\end{table*}

\emph{Quantitative Comparison:} Table~\ref{Tab:1} presents a comparison of PSNR and SSIM results for all competing methods, with the top-performing results highlighted in bold. For each noise level, we test all methods on the synthetic seismic data with four types of sizes and report the average results. The results indicate that the proposed TLSM algorithm outperforms the other competing methods in nearly all cases (the only exception is when $F = 0.1, \sigma = 0.03$ and $F = 0.1, \sigma = 0.04$ for which S2S-WTV and DNLR show slightly higher SSIM values than the proposed approach). On average, the proposed TLSM approach achieves gains of \{18.02dB, 7.43dB, 7.48dB, 10.91dB, 7.07dB, and 5.96dB\} in PSNR and \{0.2301, 0.1054, 0.1405, 0.2435, 0.1920, and 0.1173\} in SSIM compared to DRR \cite{chen2016simultaneous}, SGRDL \cite{chen2021statistics}, FR-Net \cite{qian2023unsupervised}, DNLR \cite{xu2023deep}, S2S-WTV \cite{xu2023s2s}, and UTV-LRTA \cite{qian2023improved}, respectively.

\emph{Visual Comparison:} The visual comparison results for the synthetic data are shown in Fig.~\ref{fig:1}. Due to space constraints, we only present one slice of 3-D seismic data along the crossline, inline, and time directions at the noise level $F = 0.2, \sigma = 0.01$. To highlight the performance of all competing methods, we have magnified the subregion of each denoised data for comparison. It can be observed that the proposed LTSM approach achieves superior visual results compared to the other competing methods. While the UTV-LRTA method can effectively remove the noise, it tends to introduce some blurring and may lead to the loss of fine details. These visual comparisons clearly demonstrate that the proposed TLSM approach not only effectively eliminates noise but also preserves important details. In contrast, the other competing methods either fail to completely suppress the footprint noise (\eg, DRR \cite{chen2016simultaneous}, SGRDL \cite{chen2021statistics}, DNLR \cite{xu2023deep}, and S2S-WTV \cite{xu2023s2s}) or the Gaussian noise (\eg, FR-Net \cite{qian2023unsupervised} and UTV-LRTA \cite{qian2023improved}).

\emph{Computational Efficiency:}  Computational efficiency is another crucial factor in evaluating seismic data noise suppression algorithms. To ensure a fair comparison, we report the average computational time of several competing methods on synthetic seismic data with size $100\times200\times400$ across 12 noise levels, as shown in Table~\ref{Tab:2}. The results show that the proposed TLSM algorithm achieves the fastest running time, outperforming all other competing methods, especially considering that our proposed approach was implemented solely on a CPU without leveraging GPU acceleration.

\begin{figure*}[!t]
\centering
\begin{minipage}[b]{1\linewidth}
{\includegraphics[width= 1\textwidth]{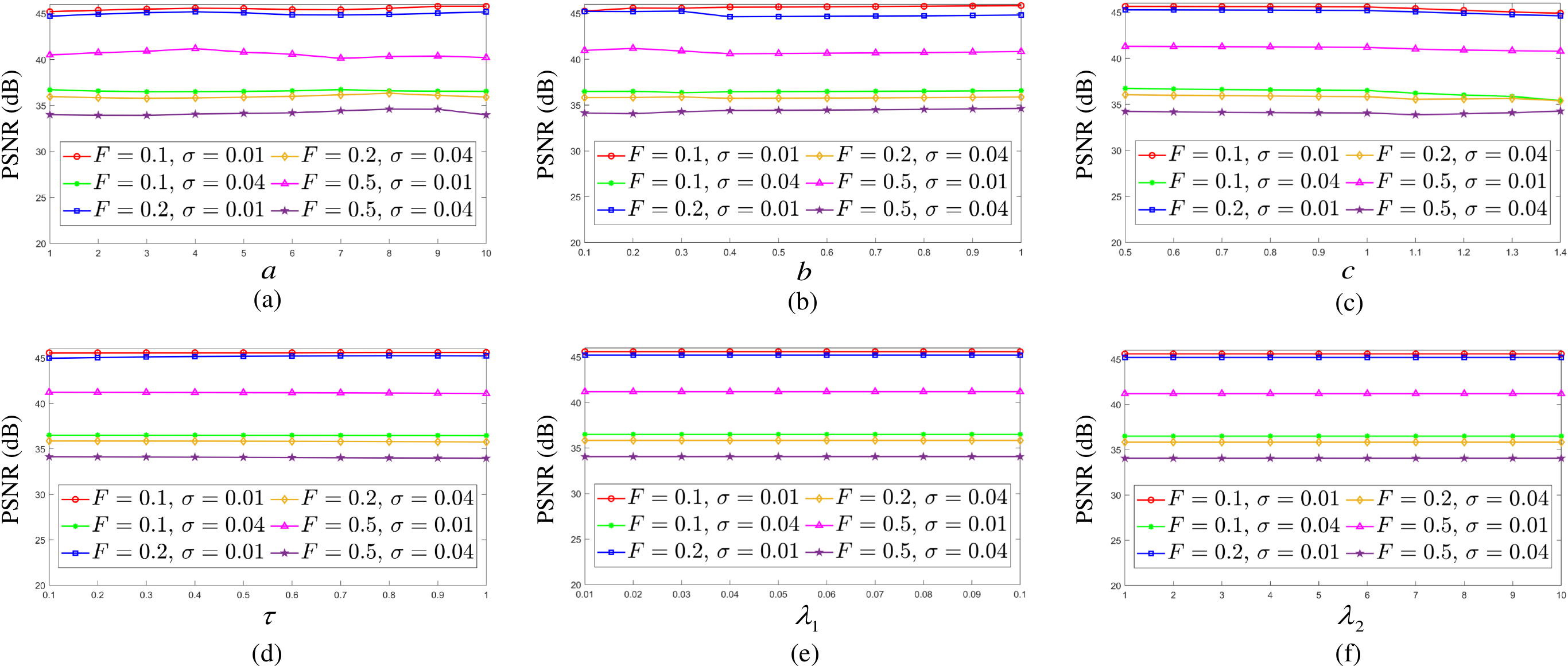}
}
\end{minipage}
\vspace{-10mm}
\caption{Parameter Analysis. (a) Analysis of different parameter $a$; (b) Analysis of different parameter $b$; (c) Analysis of different parameter $c$; (d) Analysis of different parameter $\tau$; (e) Analysis of different parameter $\lambda_1$; (f) Analysis of different parameter $\lambda_2$.}
\label{fig:4}
\vspace{-2mm}
\end{figure*}

\subsection{Field Data Experiments}
We have presented experimental results of various methods above, demonstrating the superior performance of the proposed TLSM algorithm in synthetic seismic data noise suppression. In this subsection, we extend our analysis by demonstrating the effectiveness of the TLSM algorithm on two well-known field seismic datasets. Specifically, we use a subset of the Penobscot-3D dataset \footnote{https://wiki.seg.org/wiki/Penobscot\_3D} and a subset of the Kerry-3D dataset \footnote{https://wiki.seg.org/wiki/Kerry-3D}, with sizes of $250\times150\times40$ and $400\times150\times40$, respectively.

The visual comparison results of two field seismic datasets are presented in Fig.~\ref{fig:2} and Fig.~\ref{fig:3}, respectively. Similar to the synthetic example, to effectively illustrate the noise suppression performance, the 3-D seismic data are sliced along the crossline, inline, and time directions. In the denoised results for the Penobscot-3D dataset (Fig.~\ref{fig:2}) and the Kerry-3D dataset (Fig.~\ref{fig:3}), it is evident that the proposed TLSM approach provides superior visual quality compared to the other competing methods. In comparison, other competing methods are unable to effectively remove the noise present in the image or tend to disrupt the effective seismic signals. Furthermore, to assess noise suppression quantitatively, three widely-used no-reference quality metrics are employed: BRISQUE \cite{mittal2012no}, NIQE \cite{mittal2012making}, and PIQE \cite{venkatanath2015blind}. It is important to note that lower values of BRISQUE, NIQE, and PIQE indicate better noise suppression performance. As shown in Table~\ref{Tab:25}, the proposed TLSM approach consistently achieves the lowest scores, highlighting its superior performance on the field seismic datasets compared to all other competing methods.

\begin{table}[!t]
\centering
\caption{Ablation Study: Average PSNR ($\textnormal{d}$B) comparison of TLSM-TNN, TLSM-UTV and the proposed TLSM methods for seismic data noise suppression on synthetic data with different noise levels.}
\resizebox{0.49\textwidth}{!}				
{
\scriptsize
\begin{tabular}{|cc|c|c|c|}
\hline
\multicolumn{2}{|c|}{Noise Levels}                                           & \multirow{2}{*}{TLSM-UTV} & \multirow{2}{*}{TLSM-TNN} & \multirow{2}{*}{\textbf{TLSM}} \\ \cline{1-2}
\multicolumn{1}{|l|}{Footprint}              & \multicolumn{1}{l|}{Gaussian} &                           &                           &                                \\ \hline
\multicolumn{1}{|c|}{\multirow{4}{*}{$F = 0.1$}} & $\sigma$ = 0.01                        & 43.57                     & 42.55                     & \textbf{44.06}                 \\ \cline{2-5}
\multicolumn{1}{|c|}{}                       & $\sigma$ = 0.02                        & 41.27                     & 39.13                     & \textbf{41.49}                 \\ \cline{2-5}
\multicolumn{1}{|c|}{}                       & $\sigma$ = 0.03                        & 38.06                     & 36.32                     & \textbf{38.92}                 \\ \cline{2-5}
\multicolumn{1}{|c|}{}                       & $\sigma$ = 0.04                        & 35.2                      & 34.27                     & \textbf{36.80}                 \\ \hline
\multicolumn{1}{|c|}{\multirow{4}{*}{$F = 0.2$}} & $\sigma$ = 0.01                        & 43.36                     & 42.15                     & \textbf{43.80}                 \\ \cline{2-5}
\multicolumn{1}{|c|}{}                       & $\sigma$ = 0.02                        & 39.88                     & 39.05                     & \textbf{41.36}                 \\ \cline{2-5}
\multicolumn{1}{|c|}{}                       & $\sigma$ = 0.03                        & 33.28                     & 36.16                     & \textbf{38.70}                 \\ \cline{2-5}
\multicolumn{1}{|c|}{}                       & $\sigma$ = 0.04                        & 30.79                     & 34.19                     & \textbf{36.63}                 \\ \hline
\multicolumn{1}{|c|}{\multirow{4}{*}{$F = 0.5$}} & $\sigma$ = 0.01                        & 41.45                     & 40.47                     & \textbf{41.59}                 \\ \cline{2-5}
\multicolumn{1}{|c|}{}                       & $\sigma$ = 0.02                        & 36.70                     & 38.14                     & \textbf{39.90}                 \\ \cline{2-5}
\multicolumn{1}{|c|}{}                       & $\sigma$ = 0.03                        & 33.22                     & 35.77                     & \textbf{38.04}                 \\ \cline{2-5}
\multicolumn{1}{|c|}{}                       & $\sigma$ = 0.04                        & 30.76                     & 33.95                     & \textbf{36.19}                 \\ \hline
\multicolumn{2}{|c|}{Average}                                                & 37.30                     & 37.68                     & \textbf{39.79}                 \\ \hline
\end{tabular}
}
\label{Tab:3}
\vspace{-2mm}
\end{table}

\subsection{Ablation Study}
In this section, we conduct an ablation study to further demonstrate the effectiveness of the proposed TLSM algorithm. Specifically, we compare the TLSM-UTV model (\ie, applying the LSM prior \cite{garrigues2010group} only on the UTV regularization term in Eq.~\eqref{eq:12}), TLSM-TNN (\ie, applying the LSM prior \cite{garrigues2010group} only on the TNN regularization term in Eq.~\eqref{eq:12}), and our proposed TLSM model. These three methods are tested on the synthetic seismic dataset, with parameters of the variants of Eq.~\eqref{eq:12}) tuned empirically to achieve the best noise suppression results. The average PSRN comparison results are presented in Table~\ref{Tab:3}, showing that the proposed TLSM algorithm achieves the best PSNR results than other two competing methods across 12 different noise configurations. Therefore, this ablation study demonstrates that the proposed TLSM model is effective in seismic data noise suppression.

\subsection{Algorithm Analysis}
\label{sec:5.4}
In this section, we provide some analysis for our proposed TLSM algorithm, including parameter analysis, algorithm complexity analysis, and convergence analysis.
\subsubsection{Parameter Analysis}
\label{sec:5.4.1}
Six parameters are involved in the TLSM algorithm: $a, b, c, \tau, \lambda_1,$ and $\lambda_2$. In order to analyze their individual effects, we systematically vary each parameter while keeping the others fixed. Synthetic data with sizes of $100\times200\times400$ and six different noise configurations are used for experiments, and the PSNR metric is employed to evaluate the impact of different parameter settings.

The performance comparison results for synthetic data with different parameters at various noise levels are shown in Fig.~\ref{fig:4}. It can be observed that all curves remain nearly flat, indicating that the performance of the proposed TLSM algorithm is relatively insensitive to these parameters. Based on the results of parameter analysis, we set the parameters $a, b, c, \tau, \lambda_1,$ and $\lambda_2$ to $(4, 0.2, 1, 0.5, 0.05, 1)$ for synthetic seismic data to achieve high PSNR values. Given the more complex noise patterns and diverse characteristics of field seismic data, the parameter settings are fine-tuned accordingly to ensure optimal noise suppression performance.

\begin{figure}[!t]
\centering
\begin{minipage}[b]{1\linewidth}
{\includegraphics[width= 1\textwidth]{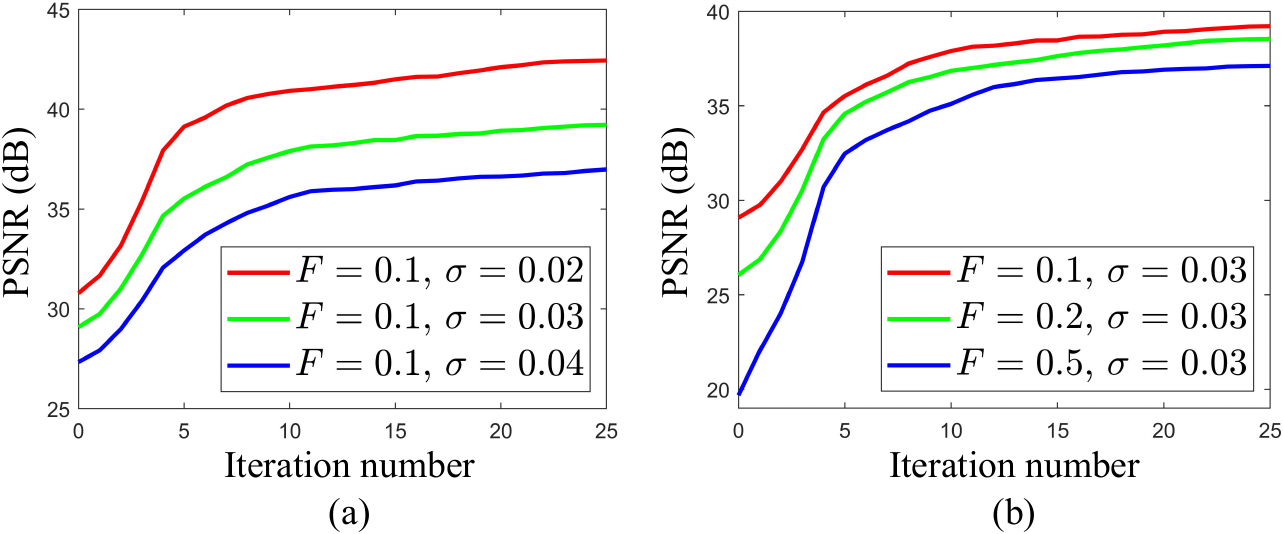}}
\end{minipage}
\vspace{-6mm}
\caption{Convergence behavior of the proposed TLSM algorithm. (a) PSNR (dB) values versus iteration numbers with $F = 0.1$. (b) PSNR (dB) values versus iteration numbers with $\sigma = 0.03$.}
\label{fig:5}
\vspace{-4mm}
\end{figure}

\subsubsection{Algorithm Complexity}
The complexity of the proposed TLSM algorithm is provided as follows. The main computational cost of the TLSM algorithm arises from two parts: 1) solving the ${\cal X}$ subproblem, which requires running the FFT operation for each of the $n_3$ matrices of size $n_1\times n_2$. Hence, the complexity of this step is $O(n_1 n_2 n_3\log(n_1 n_2))$.  2) solving the ${\cal Z}$ subproblem, which involves performing FFT and computing $n_3$ SVDs of $n_1\times n_2$ matrices. Therefore, the complexity of updating ${\cal Z}$ is $O(n_1 n_2 n_3\log(n_3) + n_{(1)} n_{(2)}^2 n_3)$, where $n_{(1)} = \text{max}(n_1,n_2)$ and $n_{(2)} = \text{min}(n_1,n_2)$. Thus, the overall per-iteration complexity of the TLSM algorithm is $O(n_1 n_2 n_3\log(n_1 n_2 n_3) + n_{(1)} n_{(2)}^2 n_3)$.

\subsubsection{Convergence Analysis}
We provide empirical evidence to characterize the convergence behavior of the proposed algorithm. Specially, we use the synthetic seismic data (size of $100\times200\times400$) as experimental cases. Fig.~\ref{fig:5} presents the curves of the PSNR values versus the number of iterations for the synthetic data across three different noise levels. It is evident that, in all cases, the PSNR curves steadily increase and eventually flatten, stabilizing as the number of iteration increases. Therefore, this experiment demonstrates that the proposed TLSM algorithm exhibits good convergence.

\section{Conclusion}
\label{sec:6}
In this paper, we have proposed a novel TLSM approach for seismic data noise suppression, which significantly differs from most existing methods that only enforce sparsity penalties through soft-thresholding or hard-thresholding operators. The proposed TLSM approach has enhanced the estimation accuracy of both the variances of the sparse tensor coefficients and the unknown sparse tensor coefficients, leading to superior performance.  Moreover, we have developed an effective ADMM algorithm to optimize the proposed seismic data noise suppression problem. Extensive experiments on both synthetic and field seismic data demonstrate that the TLSM algorithm outperforms many state-of-the-art seismic data noise suppression methods regarding quantitative and qualitative comparison metrics, while offering remarkable computational efficiency.

{\footnotesize
\bibliographystyle{IEEEtran}
\bibliography{tlsm_ref}

\begin{thebibliography}{10}
\providecommand{\url}[1]{#1}
\csname url@samestyle\endcsname
\providecommand{\newblock}{\relax}
\providecommand{\bibinfo}[2]{#2}
\providecommand{\BIBentrySTDinterwordspacing}{\spaceskip=0pt\relax}
\providecommand{\BIBentryALTinterwordstretchfactor}{4}
\providecommand{\BIBentryALTinterwordspacing}{\spaceskip=\fontdimen2\font plus
\BIBentryALTinterwordstretchfactor\fontdimen3\font minus \fontdimen4\font\relax}
\providecommand{\BIBforeignlanguage}[2]{{%
\expandafter\ifx\csname l@#1\endcsname\relax
\typeout{** WARNING: IEEEtran.bst: No hyphenation pattern has been}%
\typeout{** loaded for the language `#1'. Using the pattern for}%
\typeout{** the default language instead.}%
\else
\language=\csname l@#1\endcsname
\fi
#2}}
\providecommand{\BIBdecl}{\relax}
\BIBdecl

\bibitem{kuang2021application}
L.~Kuang, L.~He, R.~Yili, L.~Kai, S.~Mingyu, S.~Jian, and L.~Xin, ``Application and development trend of artificial intelligence in petroleum exploration and development,'' \emph{Petroleum Exploration and Development}, vol.~48, no.~1, pp. 1--14, 2021.

\bibitem{posamentier2022principles}
H.~W. Posamentier, V.~Paumard, and S.~C. Lang, ``Principles of seismic stratigraphy and seismic geomorphology i: Extracting geologic insights from seismic data,'' \emph{Earth-Science Reviews}, vol. 228, p. 103963, 2022.

\bibitem{mousavi2022deep}
S.~M. Mousavi and G.~C. Beroza, ``Deep-learning seismology,'' \emph{Science}, vol. 377, no. 6607, p. eabm4470, 2022.

\bibitem{marfurt1998suppression}
K.~J. Marfurt, R.~M. Scheet, J.~A. Sharp, and M.~G. Harper, ``Suppression of the acquisition footprint for seismic sequence attribute mapping,'' \emph{Geophysics}, vol.~63, no.~3, pp. 1024--1035, 1998.

\bibitem{al2004acquisition}
M.~Al-Bannagi, K.~Fang, P.~Kelamis, and G.~Douglass, ``Acquisition footprint suppression via the truncated svd technique,'' in \emph{SEG Technical Program Expanded Abstracts 2004}.\hskip 1em plus 0.5em minus 0.4em\relax Society of Exploration Geophysicists, 2004, pp. 1957--1960.

\bibitem{falconer2008attribute}
S.~Falconer and K.~J. Marfurt, ``Attribute-driven footprint suppression,'' in \emph{SEG International Exposition and Annual Meeting}.\hskip 1em plus 0.5em minus 0.4em\relax SEG, 2008, pp. SEG--2008.

\bibitem{han2022gaussian}
J.~Han, Q.~L{\"u}, B.~Gu, and Z.~Xing, ``Gaussian beam summation migration of deep reflection seismic data: Numerical examples,'' \emph{IEEE Geoscience and Remote Sensing Letters}, vol.~19, pp. 1--5, 2022.

\bibitem{zhang2021deep}
W.~Zhang and J.~Gao, ``Deep-learning full-waveform inversion using seismic migration images,'' \emph{IEEE Transactions on Geoscience and Remote Sensing}, vol.~60, pp. 1--18, 2021.

\bibitem{gu2023seismic}
X.~Gu, W.~Lu, Y.~Ao, Y.~Li, and C.~Song, ``Seismic stratigraphic interpretation based on deep active learning,'' \emph{IEEE Transactions on Geoscience and Remote Sensing}, vol.~61, pp. 1--11, 2023.

\bibitem{abma1995lateral}
R.~Abma and J.~Claerbout, ``Lateral prediction for noise attenuation by tx and fx techniques,'' \emph{Geophysics}, vol.~60, no.~6, pp. 1887--1896, 1995.

\bibitem{zhang2009footprint}
R.~Zhang, ``Footprint suppression with basis pursuit denoising,'' in \emph{SEG International Exposition and Annual Meeting}.\hskip 1em plus 0.5em minus 0.4em\relax SEG, 2009, pp. SEG--2009.

\bibitem{liu2012spatiotemporal}
Y.~Liu, Y.~Li, P.~Nie, and Q.~Zeng, ``Spatiotemporal time--frequency peak filtering method for seismic random noise reduction,'' \emph{IEEE Geoscience and Remote Sensing Letters}, vol.~10, no.~4, pp. 756--760, 2012.

\bibitem{li2017multidimensional}
C.~Li, G.~Liu, Z.~Hao, S.~Zu, F.~Mi, and X.~Chen, ``Multidimensional seismic data reconstruction using frequency-domain adaptive prediction-error filter,'' \emph{IEEE Transactions on Geoscience and Remote Sensing}, vol.~56, no.~4, pp. 2328--2336, 2017.

\bibitem{cvetkovic20072d}
M.~Cvetkovic, S.~Falconer, K.~J. Marfurt, and S.~Ch{\'a}vez-P{\'e}rez, ``2d stationary wavelet transform based acquisition footprint suppression,'' in \emph{SEG International Exposition and Annual Meeting}.\hskip 1em plus 0.5em minus 0.4em\relax SEG, 2007, pp. SEG--2007.

\bibitem{alali2018attribute}
A.~Alali, G.~Machado, and K.~J. Marfurt, ``Attribute-assisted footprint suppression using a 2d continuous wavelet transform,'' \emph{Interpretation}, vol.~6, no.~2, pp. T457--T470, 2018.

\bibitem{wang2019efficient}
B.~Wang and J.~Geng, ``Efficient deblending in the pfk domain based on compressive sensing,'' \emph{IEEE Transactions on Geoscience and Remote Sensing}, vol.~58, no.~2, pp. 995--1003, 2019.

\bibitem{zhang2018multicomponent}
C.~Zhang and M.~van~der Baan, ``Multicomponent microseismic data denoising by 3d shearlet transform,'' \emph{Geophysics}, vol.~83, no.~3, pp. A45--A51, 2018.

\bibitem{gomez2020footprint}
J.~L. G{\'o}mez and D.~R. Velis, ``Footprint removal from seismic data with residual dictionary learning,'' \emph{Geophysics}, vol.~85, no.~4, pp. V355--V365, 2020.

\bibitem{liu2021dictionary}
D.~Liu, L.~Gao, X.~Wang, and W.~Chen, ``A dictionary learning method with atom splitting for seismic footprint suppression,'' \emph{Geophysics}, vol.~86, no.~6, pp. V509--V523, 2021.

\bibitem{zhou2023coherent}
Z.~Zhou, M.~Bai, J.~Wu, and Y.~Cui, ``Coherent noise attenuation by kurtosis-guided adaptive dictionary learning based on variational sparse representation,'' \emph{IEEE Transactions on Geoscience and Remote Sensing}, vol.~61, pp. 1--10, 2023.

\bibitem{chen2021statistics}
W.~Chen, O.~M. Saad, H.~Wang, and Y.~Chen, ``Statistics-guided residual dictionary learning for footprint noise removal,'' \emph{IEEE Transactions on Geoscience and Remote Sensing}, vol.~60, pp. 1--11, 2021.

\bibitem{chen2016simultaneous}
Y.~Chen, D.~Zhang, Z.~Jin, X.~Chen, S.~Zu, W.~Huang, and S.~Gan, ``Simultaneous denoising and reconstruction of 5-d seismic data via damped rank-reduction method,'' \emph{Geophysical Journal International}, vol. 206, no.~3, pp. 1695--1717, 2016.

\bibitem{oboue2021enhanced}
Y.~A. S.~I. Obou{\'e} and Y.~Chen, ``Enhanced low-rank matrix estimation for simultaneous denoising and reconstruction of 5d seismic data,'' \emph{Geophysics}, vol.~86, no.~5, pp. V459--V470, 2021.

\bibitem{niu2021seismic}
X.~Niu, L.~Fu, W.~Zhang, and Y.~Li, ``Seismic data interpolation based on simultaneously sparse and low-rank matrix recovery,'' \emph{IEEE Transactions on Geoscience and Remote Sensing}, vol.~60, pp. 1--13, 2021.

\bibitem{lin2023structure}
P.~Lin, S.~Peng, Y.~Xiang, C.~Li, X.~Cui, and W.~Zhang, ``Structure-oriented cur low-rank approximation for random noise attenuation of seismic data,'' \emph{IEEE Transactions on Geoscience and Remote Sensing}, 2023.

\bibitem{wang2023seismic}
C.~Wang, Z.~Gu, and Z.~Zhu, ``Seismic data reconstruction and denoising by enhanced hankel low-rank matrix estimation,'' \emph{IEEE Transactions on Geoscience and Remote Sensing}, vol.~61, pp. 1--13, 2023.

\bibitem{goldfarb2014robust}
D.~Goldfarb and Z.~Qin, ``Robust low-rank tensor recovery: Models and algorithms,'' \emph{SIAM Journal on Matrix Analysis and Applications}, vol.~35, no.~1, pp. 225--253, 2014.

\bibitem{chen2018tensor}
Y.~Chen, S.~Wang, and Y.~Zhou, ``Tensor nuclear norm-based low-rank approximation with total variation regularization,'' \emph{IEEE Journal of Selected Topics in Signal Processing}, vol.~12, no.~6, pp. 1364--1377, 2018.

\bibitem{qian2023unsupervised}
F.~Qian, Y.~Yue, Y.~He, H.~Yu, Y.~Zhou, J.~Tang, and G.~Hu, ``Unsupervised seismic footprint removal with physical prior augmented deep autoencoder,'' \emph{IEEE Transactions on Geoscience and Remote Sensing}, vol.~61, pp. 1--20, 2023.

\bibitem{qian2023improved}
F.~Qian, Y.~He, Y.~Yue, Y.~Zhou, B.~Wu, and G.~Hu, ``Improved low-rank tensor approximation for seismic random plus footprint noise suppression,'' \emph{IEEE Transactions on Geoscience and Remote Sensing}, vol.~61, pp. 1--19, 2023.

\bibitem{feng2021low}
J.~Feng, X.~Liu, X.~Li, W.~Xu, and B.~Liu, ``Low-rank tensor minimization method for seismic denoising based on variational mode decomposition,'' \emph{IEEE Geoscience and Remote Sensing Letters}, vol.~19, pp. 1--5, 2021.

\bibitem{feng2021seismic}
J.~Feng, X.~Li, X.~Liu, C.~Chen, and H.~Chen, ``Seismic data denoising based on tensor decomposition with total variation,'' \emph{IEEE Geoscience and Remote Sensing Letters}, vol.~18, no.~7, pp. 1303--1307, 2021.

\bibitem{liu2024simultaneous}
L.~Liu and Y.~Li, ``Simultaneous sparse and low rank regularization for seismic data denoising,'' \emph{IEEE Geoscience and Remote Sensing Letters}, 2024.

\bibitem{lu2016tensor}
C.~Lu, J.~Feng, Y.~Chen, W.~Liu, Z.~Lin, and S.~Yan, ``Tensor robust principal component analysis: Exact recovery of corrupted low-rank tensors via convex optimization,'' in \emph{Proceedings of the IEEE conference on computer vision and pattern recognition}, 2016, pp. 5249--5257.

\bibitem{wang2015improved}
B.~Wang, X.~Chen, J.~Li, and J.~Cao, ``An improved weighted projection onto convex sets method for seismic data interpolation and denoising,'' \emph{IEEE Journal of Selected Topics in Applied Earth Observations and Remote Sensing}, vol.~9, no.~1, pp. 228--235, 2015.

\bibitem{li2022simultaneous}
C.~Li, X.~Wen, X.~Liu, and S.~Zu, ``Simultaneous seismic data interpolation and denoising based on nonsubsampled contourlet transform integrating with two-step iterative log thresholding algorithm,'' \emph{IEEE Transactions on Geoscience and Remote Sensing}, vol.~60, pp. 1--10, 2022.

\bibitem{osher2005iterative}
S.~Osher, M.~Burger, D.~Goldfarb, J.~Xu, and W.~Yin, ``An iterative regularization method for total variation-based image restoration,'' \emph{Multiscale Modeling \& Simulation}, vol.~4, no.~2, pp. 460--489, 2005.

\bibitem{aharon2006k}
M.~Aharon, M.~Elad, and A.~Bruckstein, ``K-svd: An algorithm for designing overcomplete dictionaries for sparse representation,'' \emph{IEEE Transactions on Signal Processing}, vol.~54, no.~11, pp. 4311--4322, 2006.

\bibitem{cai2010singular}
J.-F. Cai, E.~J. Cand{\`e}s, and Z.~Shen, ``A singular value thresholding algorithm for matrix completion,'' \emph{SIAM Journal on Optimization}, vol.~20, no.~4, pp. 1956--1982, 2010.

\bibitem{cheng2015application}
J.~Cheng, K.~Chen, and M.~D. Sacchi, ``Application of robust principal component analysis (rpca) to suppress erratic noise in seismic records,'' in \emph{SEG Technical Program Expanded Abstracts 2015}.\hskip 1em plus 0.5em minus 0.4em\relax Society of Exploration Geophysicists, 2015, pp. 4646--4651.

\bibitem{carroll1970analysis}
J.~D. Carroll and J.-J. Chang, ``Analysis of individual differences in multidimensional scaling via an n-way generalization of “eckart-young” decomposition,'' \emph{Psychometrika}, vol.~35, no.~3, pp. 283--319, 1970.

\bibitem{zhu2019seismic}
W.~Zhu, S.~M. Mousavi, and G.~C. Beroza, ``Seismic signal denoising and decomposition using deep neural networks,'' \emph{IEEE Transactions on Geoscience and Remote Sensing}, vol.~57, no.~11, pp. 9476--9488, 2019.

\bibitem{wang2020generative}
H.~Wang, Y.~Li, and X.~Dong, ``Generative adversarial network for desert seismic data denoising,'' \emph{IEEE Transactions on Geoscience and Remote Sensing}, vol.~59, no.~8, pp. 7062--7075, 2020.

\bibitem{wang2024efgw}
X.~Wang, J.~Ma, X.~Dong, T.~Zhong, and S.~Dong, ``Efgw-unet: A deep-learning-based approach for weak signal recovery in seismic data,'' \emph{IEEE Transactions on Geoscience and Remote Sensing}, 2024.

\bibitem{wang2022learning}
F.~Wang, B.~Yang, Y.~Wang, and M.~Wang, ``Learning from noisy data: An unsupervised random denoising method for seismic data using model-based deep learning,'' \emph{IEEE Transactions on Geoscience and Remote Sensing}, vol.~60, pp. 1--14, 2022.

\bibitem{xu2023deep}
Z.~Xu, Y.~Luo, B.~Wu, D.~Meng, and Y.~Chen, ``Deep nonlocal regularizer: A self-supervised learning method for 3d seismic denoising,'' \emph{IEEE Transactions on Geoscience and Remote Sensing}, 2023.

\bibitem{xu2023s2s}
Z.~Xu, Y.~Luo, B.~Wu, and D.~Meng, ``S2s-wtv: Seismic data noise attenuation using weighted total variation regularized self-supervised learning,'' \emph{IEEE Transactions on Geoscience and Remote Sensing}, vol.~61, pp. 1--15, 2023.

\bibitem{lecun2015deep}
Y.~LeCun, Y.~Bengio, and G.~Hinton, ``Deep learning,'' \emph{nature}, vol. 521, no. 7553, pp. 436--444, 2015.

\bibitem{goodfellow2020generative}
I.~Goodfellow, J.~Pouget-Abadie, M.~Mirza, B.~Xu, D.~Warde-Farley, S.~Ozair, A.~Courville, and Y.~Bengio, ``Generative adversarial networks,'' \emph{Communications of the ACM}, vol.~63, no.~11, pp. 139--144, 2020.

\bibitem{ronneberger2015u}
O.~Ronneberger, P.~Fischer, and T.~Brox, ``U-net: Convolutional networks for biomedical image segmentation,'' in \emph{Medical image computing and computer-assisted intervention--MICCAI 2015: 18th international conference, Munich, Germany, October 5-9, 2015, proceedings, part III 18}.\hskip 1em plus 0.5em minus 0.4em\relax Springer, 2015, pp. 234--241.

\bibitem{liu2021unsupervised}
B.~Liu, J.~Yue, Z.~Zuo, X.~Xu, C.~Fu, S.~Yang, and P.~Jiang, ``Unsupervised deep learning for random noise attenuation of seismic data,'' \emph{IEEE Geoscience and Remote Sensing Letters}, vol.~19, pp. 1--5, 2021.

\bibitem{qiu2021deep}
C.~Qiu, B.~Wu, N.~Liu, X.~Zhu, and H.~Ren, ``Deep learning prior model for unsupervised seismic data random noise attenuation,'' \emph{IEEE Geoscience and Remote Sensing Letters}, vol.~19, pp. 1--5, 2021.

\bibitem{meng2021self}
F.~Meng, Q.~Fan, and Y.~Li, ``Self-supervised learning for seismic data reconstruction and denoising,'' \emph{IEEE Geoscience and Remote Sensing Letters}, vol.~19, pp. 1--5, 2021.

\bibitem{quan2020self2self}
Y.~Quan, M.~Chen, T.~Pang, and H.~Ji, ``Self2self with dropout: Learning self-supervised denoising from single image,'' in \emph{Proceedings of the IEEE/CVF conference on computer vision and pattern recognition}, 2020, pp. 1890--1898.

\bibitem{meng2024stochastic}
C.~Meng, J.~Gao, Y.~Tian, H.~Chen, W.~Zhang, and R.~Luo, ``Stochastic solutions for simultaneous seismic data denoising and reconstruction via score-based generative models,'' \emph{IEEE Transactions on Geoscience and Remote Sensing}, 2024.

\bibitem{wang2020structure}
D.~Wang, J.~Gao, N.~Liu, and X.~Jiang, ``Structure-oriented dtgv regularization for random noise attenuation in seismic data,'' \emph{IEEE Transactions on Geoscience and Remote Sensing}, vol.~59, no.~2, pp. 1757--1771, 2020.

\bibitem{liu2021high}
X.~Liu, Q.~Li, C.~Yuan, J.~Li, X.~Chen, and Y.~Chen, ``High-order directional total variation for seismic noise attenuation,'' \emph{IEEE Transactions on Geoscience and Remote Sensing}, vol.~60, pp. 1--13, 2021.

\bibitem{zha2020image}
Z.~Zha, X.~Yuan, J.~Zhou, C.~Zhu, and B.~Wen, ``Image restoration via simultaneous nonlocal self-similarity priors,'' \emph{IEEE Transactions on Image Processing}, vol.~29, pp. 8561--8576, 2020.

\bibitem{huang2017mixed}
T.~Huang, W.~Dong, X.~Xie, G.~Shi, and X.~Bai, ``Mixed noise removal via laplacian scale mixture modeling and nonlocal low-rank approximation,'' \emph{IEEE Transactions on Image Processing}, vol.~26, no.~7, pp. 3171--3186, 2017.

\bibitem{dong2015image}
W.~Dong, G.~Shi, Y.~Ma, and X.~Li, ``Image restoration via simultaneous sparse coding: Where structured sparsity meets gaussian scale mixture,'' \emph{International Journal of Computer Vision}, vol. 114, pp. 217--232, 2015.

\bibitem{zha2020benchmark}
Z.~Zha, X.~Yuan, B.~Wen, J.~Zhou, J.~Zhang, and C.~Zhu, ``A benchmark for sparse coding: When group sparsity meets rank minimization,'' \emph{IEEE Transactions on Image Processing}, vol.~29, pp. 5094--5109, 2020.

\bibitem{dong2015low}
W.~Dong, G.~Li, G.~Shi, X.~Li, and Y.~Ma, ``Low-rank tensor approximation with laplacian scale mixture modeling for multiframe image denoising,'' in \emph{Proceedings of the IEEE International Conference on Computer Vision}, 2015, pp. 442--449.

\bibitem{xue2022laplacian}
J.~Xue, Y.~Zhao, Y.~Bu, J.~C.-W. Chan, and S.~G. Kong, ``When laplacian scale mixture meets three-layer transform: A parametric tensor sparsity for tensor completion,'' \emph{IEEE Transactions on Cybernetics}, vol.~52, no.~12, pp. 13\,887--13\,901, 2022.

\bibitem{box2011bayesian}
G.~E. Box and G.~C. Tiao, \emph{Bayesian inference in statistical analysis}.\hskip 1em plus 0.5em minus 0.4em\relax John Wiley \& Sons, 2011.

\bibitem{garrigues2010group}
P.~Garrigues and B.~Olshausen, ``Group sparse coding with a laplacian scale mixture prior,'' \emph{Advances in neural information processing systems}, vol.~23, 2010.

\bibitem{boyd2011distributed}
S.~Boyd, N.~Parikh, E.~Chu, B.~Peleato, J.~Eckstein \emph{et~al.}, ``Distributed optimization and statistical learning via the alternating direction method of multipliers,'' \emph{Foundations and Trends{\textregistered} in Machine learning}, vol.~3, no.~1, pp. 1--122, 2011.

\bibitem{chang2015anisotropic}
Y.~Chang, L.~Yan, H.~Fang, and C.~Luo, ``Anisotropic spectral-spatial total variation model for multispectral remote sensing image destriping,'' \emph{IEEE Transactions on Image Processing}, vol.~24, no.~6, pp. 1852--1866, 2015.

\bibitem{chan2013constrained}
R.~H. Chan, M.~Tao, and X.~Yuan, ``Constrained total variation deblurring models and fast algorithms based on alternating direction method of multipliers,'' \emph{SIAM Journal on Imaging Sciences}, vol.~6, no.~1, pp. 680--697, 2013.

\bibitem{zhang2014novel}
Z.~Zhang, G.~Ely, S.~Aeron, N.~Hao, and M.~Kilmer, ``Novel methods for multilinear data completion and de-noising based on tensor-svd,'' in \emph{Proceedings of the IEEE conference on computer vision and pattern recognition}, 2014, pp. 3842--3849.

\bibitem{kilmer2011factorization}
M.~E. Kilmer and C.~D. Martin, ``Factorization strategies for third-order tensors,'' \emph{Linear Algebra and its Applications}, vol. 435, no.~3, pp. 641--658, 2011.

\bibitem{donoho1995noising}
D.~L. Donoho, ``De-noising by soft-thresholding,'' \emph{IEEE Transactions on Information Theory}, vol.~41, no.~3, pp. 613--627, 1995.

\bibitem{wang2004image}
Z.~Wang, A.~C. Bovik, H.~R. Sheikh, and E.~P. Simoncelli, ``Image quality assessment: from error visibility to structural similarity,'' \emph{IEEE Transactions on Image Processing}, vol.~13, no.~4, pp. 600--612, 2004.

\bibitem{mittal2012no}
A.~Mittal, A.~K. Moorthy, and A.~C. Bovik, ``No-reference image quality assessment in the spatial domain,'' \emph{IEEE Transactions on Image Processing}, vol.~21, no.~12, pp. 4695--4708, 2012.

\bibitem{mittal2012making}
A.~Mittal, R.~Soundararajan, and A.~C. Bovik, ``Making a “completely blind” image quality analyzer,'' \emph{IEEE Signal Processing Letters}, vol.~20, no.~3, pp. 209--212, 2012.

\bibitem{venkatanath2015blind}
N.~Venkatanath, D.~Praneeth, M.~C. Bh, S.~S. Channappayya, and S.~S. Medasani, ``Blind image quality evaluation using perception based features,'' in \emph{2015 twenty first national conference on communications (NCC)}.\hskip 1em plus 0.5em minus 0.4em\relax IEEE, 2015, pp. 1--6.

\end{thebibliography}
}

\vspace{-10mm}
\begin{IEEEbiography}[{\includegraphics[width=1in,height=1.25in,clip,keepaspectratio]{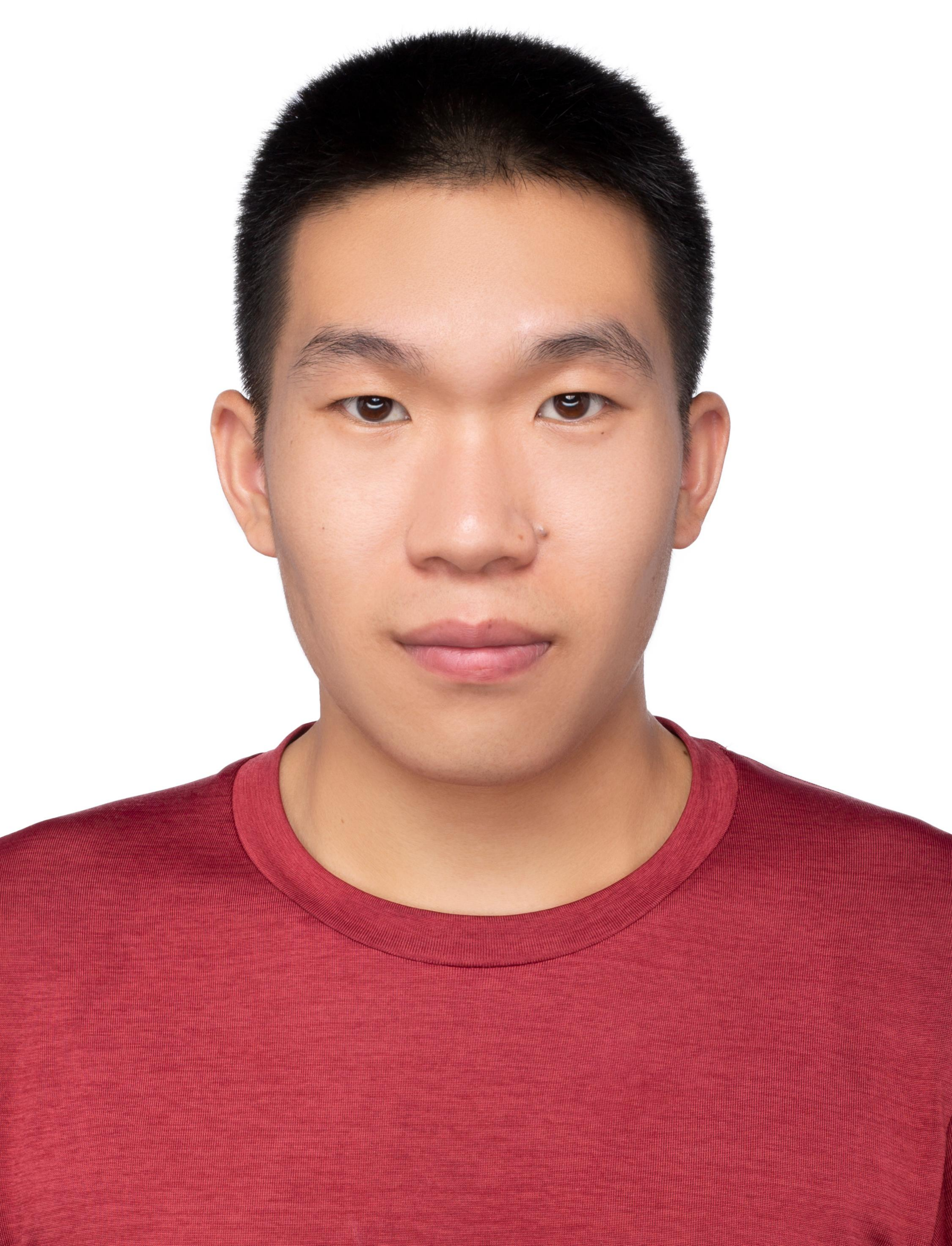}}]{Sirui Pan}
received the B.S. degree from the School of Information and Communication Engineering, University of Electronic Science and Technology of China, Chengdu, China, in 2020. He is currently pursuing the M.S. degree at the College of Communication Engineering, Jilin University, Changchun, China. His research interests include seismic data processing.
\end{IEEEbiography}
\vspace{-10 mm}
\begin{IEEEbiography}[{\includegraphics[width=1in,height=1.25in,clip,keepaspectratio]{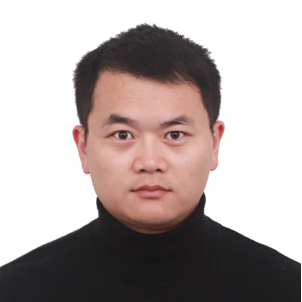}}]{Zhiyuan Zha} (Senior Member, IEEE)
received the Ph.D. degree with the School of Electronic Science and Engineering, Nanjing University, Nanjing, China, in 2018. He is a Professor at the College of Communication Engineering, Jilin University. Prior to that, he was a Senior Research Fellow with Nanyang Technological University, Singapore. His current research interests include inverse problems in image/video processing, sparse signal representation and machine learning. He was a recipient of the  Platinum Best Paper Award and  the Best Paper Runner Up Award at the IEEE International Conference on Multimedia and Expo  in 2017 and 2020, respectively. He has headed the special issue ``High-dimensional imaging: emerging challenges and advances in reconstruction and restoration" as the Lead Guest Editor in the IEEE JOURNAL OF SELECTED TOPICS IN SIGNAL PROCESSING in 2025. He has been an Associate Editor for IEEE TRANSACTIONS ON IMAGE PROCESSING since 2024, a Handling Editor for SIGNAL PROCESSING since 2024, and an Associate Editor for THE VISUAL COMPUTER since 2023. He was selected for the National High-Level Youth Talent Program, China, in 2024.
\end{IEEEbiography}
\vspace{-10 mm}
\begin{IEEEbiography}[{\includegraphics[width=1in,height=1.25in,clip,keepaspectratio]{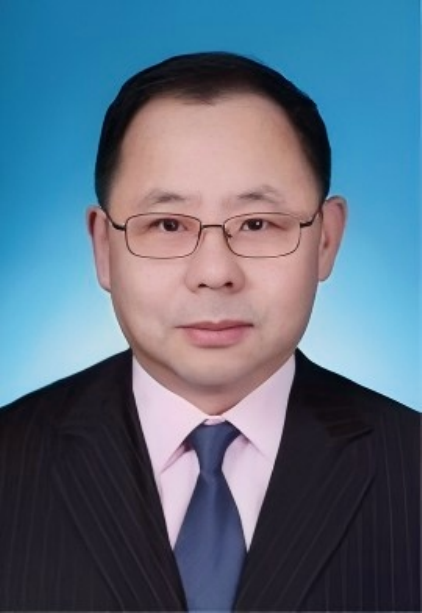}}]{Shigang Wang} (Member, IEEE)
received the B.S. degree from Northeastern University in 1983, the M.S. degree in communication and electronic from Jilin University of Technology in 1998, and the Ph.D. degree in communication and information system from Jilin University in 2001. He is currently a Professor with the College of Communication Engineering, Jilin University. His research interests include image and video coding and multidimensional signal processing.
\end{IEEEbiography}
\vspace{-10 mm}
\begin{IEEEbiography}[{\includegraphics[width=1in,height=1.25in,clip,keepaspectratio]{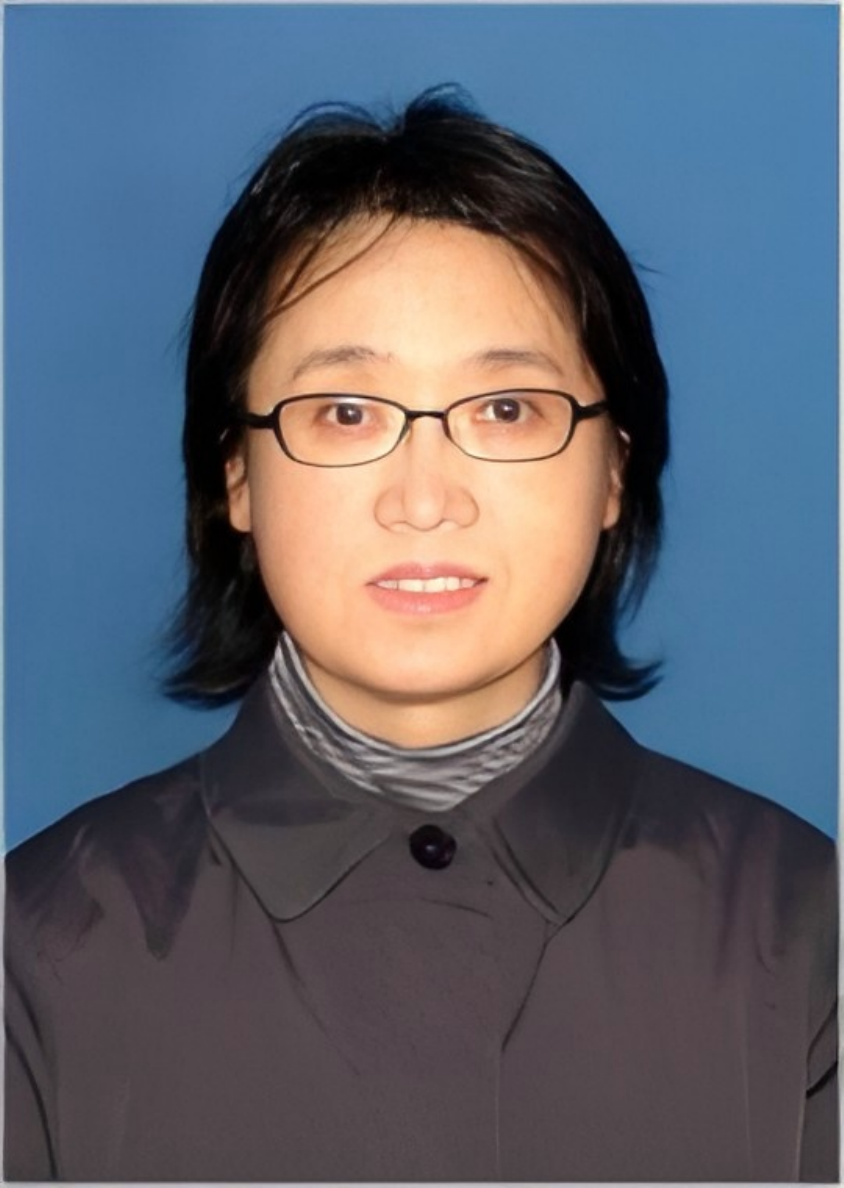}}]{Yue Li} (Senior Member, IEEE)
received the B.S. degree in automatic control from Changchun University of Technology, Changchun, China, in 1982, and the Ph.D. degree in earth scope from Jilin University, Changchun, in 2001.

She is currently a Professor of Communication and Information Systems with the Department of Communication Engineering, Jilin University. She has authored more than 100 articles on deep learning, weak signal detection, real-time signal processing, nonlinear dynamic systems, random signal processing, the Radon-Wigner transform, particle filters, time-frequency analysis, weak seismic information extraction, and medical signal processing. She has coauthored two books, METHODOLOGY OF PERIODIC OSCILLATOR DETECTION and PERIODIC OSCILLATOR SYSTEM AND DETECTION. She was approved by several general and key programs of the National Natural Science Foundation of China.
\end{IEEEbiography}
\vspace{-10 mm}
\begin{IEEEbiography}[{\includegraphics[width=1in,height=1.25in,clip,keepaspectratio]{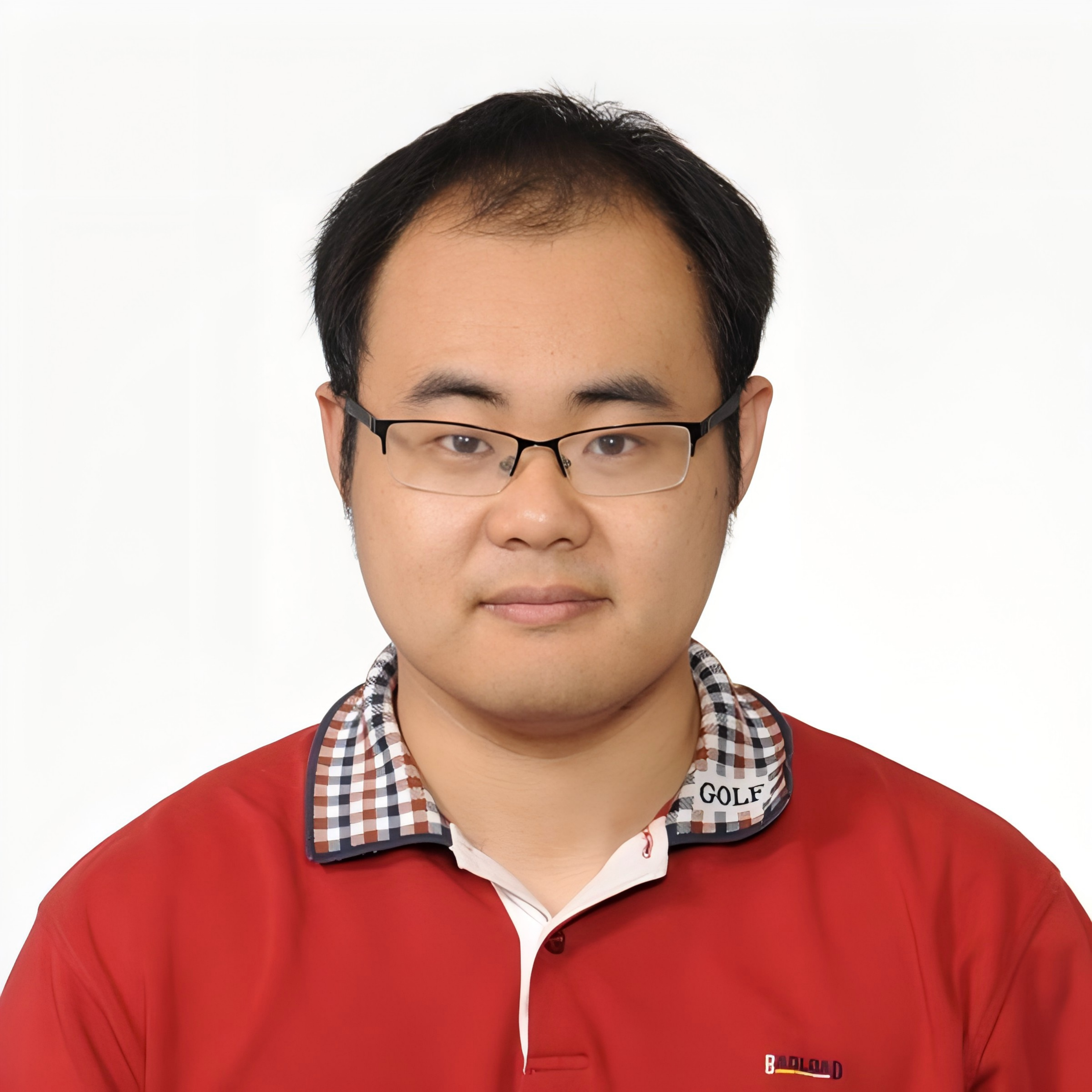}}]{Zipei Fan} (Member, IEEE)
received his B.S. degree in Computer Science from Beihang University, China, in 2012, both M.S. and a Ph.D. degree in Civil Engineering from The University of Tokyo, Japan, in 2014 and 2017 respectively. He became Project Researcher and Project Assistant Professor in
2017 and 2019, and he has promoted to Project Lecturer at the Center for Spatial Information Science, the University of Tokyo in 2020. His
research interests include ubiquitous computing, machine learning, spatio-temporal data mining, and heterogeneous data fusion. He was selected for the National High-Level Youth Talent Program, China, in 2023.
\end{IEEEbiography}
\vspace{-10 mm}
\begin{IEEEbiography}[{\includegraphics[width=1in,height=1.25in,clip,keepaspectratio]{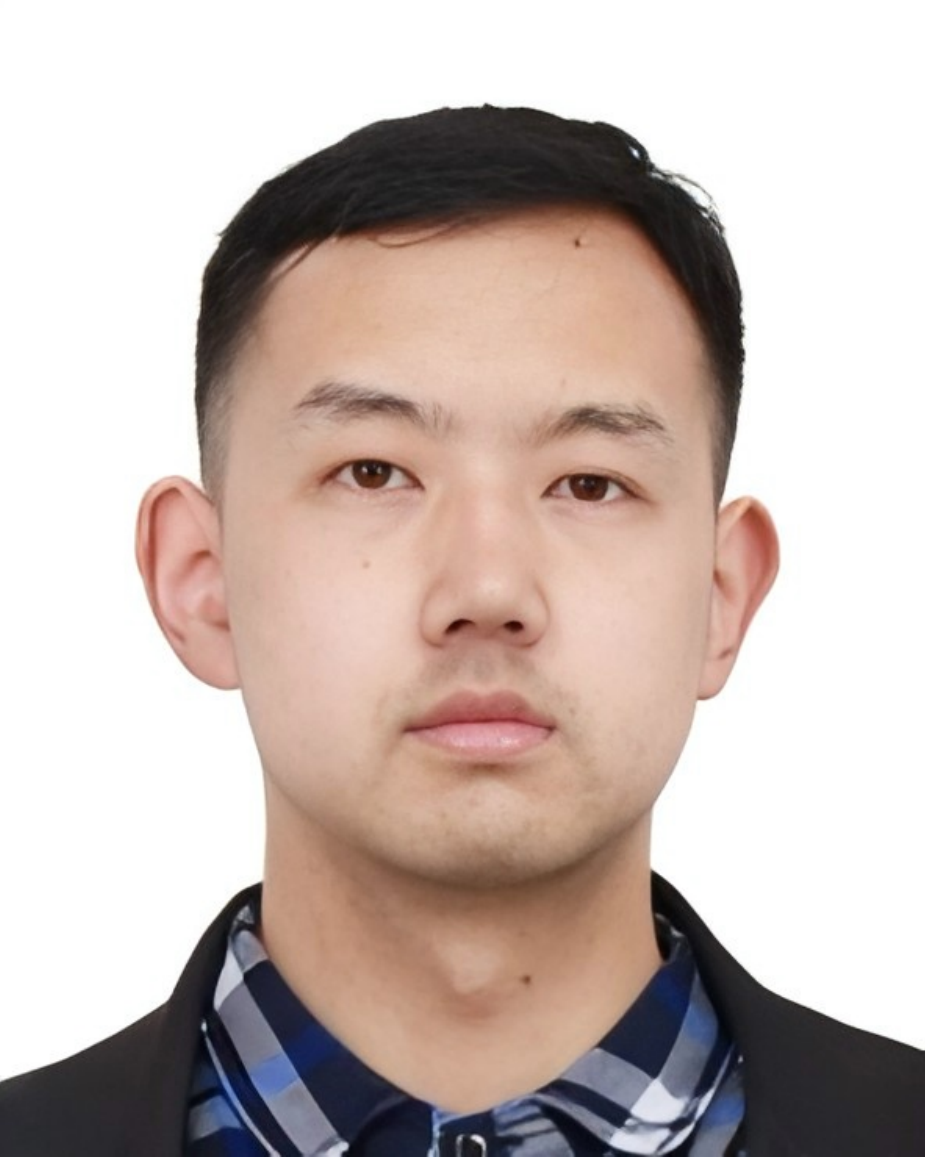}}]{Gang Yan} (Member, IEEE)
received his Ph.D. degree in Electrical and Computer Engineering from the State University of New York at Binghamton in 2023. Prior to that, He earned an M.S. (2019) and a B.S. (2016) in Statistics from Nankai University (China). In 2024, he served as a postdoctoral scholar at the University of California, Merced. He is currently a full professor with the College of Computer Science and Technology, Jilin University. His research centers on high-performance network optimization, cloud and edge computing, distributed machine learning, and federated learning. He was selected for the National High-Level Youth Talent Program, China, in 2024.
\end{IEEEbiography}
\vspace{-10 mm}
\begin{IEEEbiography}[{\includegraphics[width=1in,height=1.25in,clip,keepaspectratio]{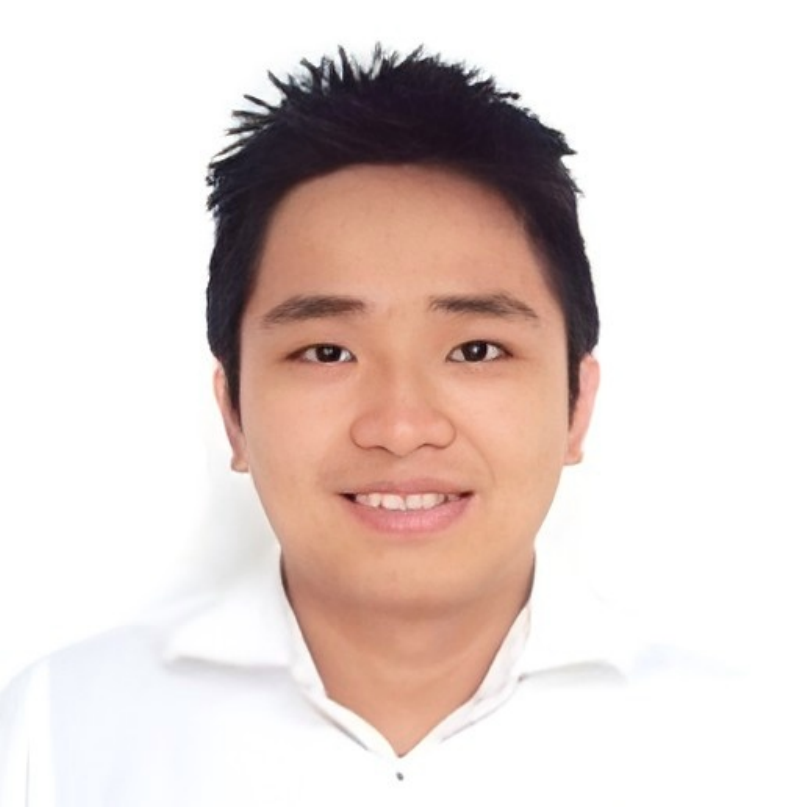}}]{Binh T. Nguyen} (Member, IEEE)
received the Ph.D. degree (Hons.) from Ecole Polytechnique, Paris, France, in 2012. He is currently the Head of the Department of Computer Science and an Associate Professor of computer science with the Faculty of Mathematics and Computer Science, University of Science, Vietnam National University Ho Chi Minh City. He has over ten years of experience in AI and data science. Up to now, he has had over 100 publications and four patents filed in USA and Canada. He has substantial experience building research and development teams to help companies or startups deliver AI products.
\end{IEEEbiography}
\vspace{-10 mm}
\begin{IEEEbiography}[{\includegraphics[width=1in,height=1.25in,clip,keepaspectratio]{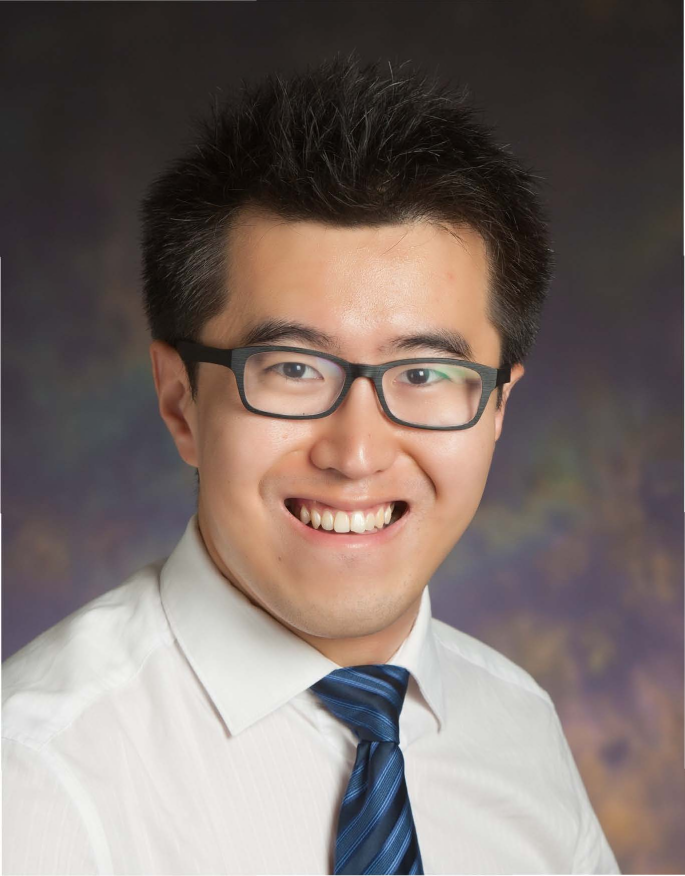}}]{Bihan Wen} (Senior Member, IEEE)
received the B.Eng. degree in electrical and electronic engineering from Nanyang Technological University, Singapore, in 2012, the MS and PhD degrees in electrical and computer engineering from University of Illinois at Urbana-Champaign, USA, in 2015 and 2018, respectively. He is currently a Nanyang Assistant Professor with the School of Electrical and Electronic Engineering at Nanyang Technological University, Singapore.

He was the recipient of the 2016 Yee Fellowship, and the 2012 Professional Engineers Board Gold Medal. His research interests span areas of machine learning, computational imaging, computer vision, image and video processing, and big data applications. He was a recipient of the Best Paper Runner Up Award at the IEEE International Conference on Multimedia and Expo in 2020. He has been the Associate Editor of IEEE TRANSACTIONS ON CIRCUITS AND SYSTEMS FOR VIDEO TECHNOLOGY since 2022.
\end{IEEEbiography}

\vspace{-10 mm}
\begin{IEEEbiography}[{\includegraphics[width=1in,height=1.25in,clip,keepaspectratio]{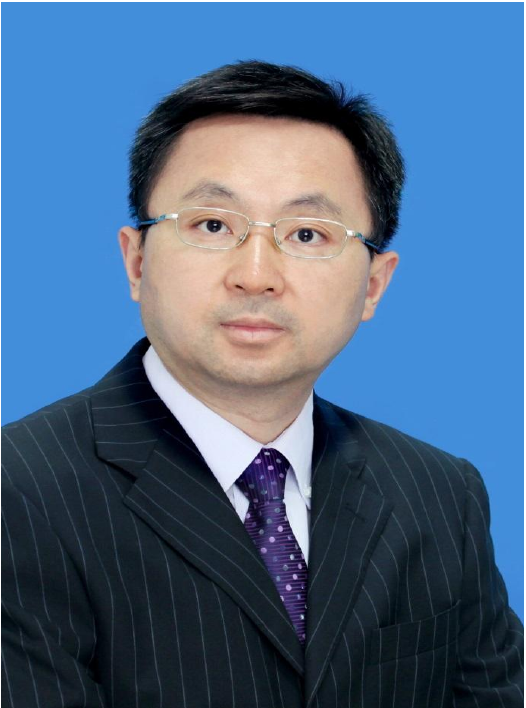}}]{Ce Zhu} (Fellow, IEEE)
received the B.S. degree in electronic and information engineering from Sichuan University, Chengdu, China, in 1989, and the M.Eng. and Ph.D. degrees in electronic and information engineering from Southeast University, Nanjing, China, in 1992 and 1994, respectively. He was a Postdoctoral Researcher with the Chinese University of Hong Kong in 1995, the City University of Hong Kong, and the University of Melbourne, Australia, from 1996 to 1998. From 1998 to 2012, he was with Nanyang Technological University, Singapore, where he was a Research Fellow, a Program Manager, an Assistant Professor, and then promoted to an Associate Professor in 2005. He has been with the University of Electronic Science and Technology of China (UESTC), Chengdu, China, as a Professor since 2012, and is the Dean of Glasgow College, a joint school between the University of Glasgow, Glasgow, U.K. and UESTC. His research interests include video coding and communications, video analysis and processing, 3-D video, and visual perception and applications. He was the co-recipient of multiple paper awards at international conferences, including the most recent Best Demo Award at IEEE MMSP 2022, and the Best Paper Runner Up Award at IEEE ICME 2020. He has served on the editorial boards of a few journals, including as an Associate Editor for IEEE TRANSACTIONS ON IMAGE PROCESSING, IEEE TRANSACTIONS ON CIRCUITS AND SYSTEMS FOR VIDEO TECHNOLOGY, IEEE TRANSACTIONS ON BROADCASTING, and IEEE SIGNAL PROCESSING LETTERS, the Editor for IEEE COMMUNICATIONS SURVEYS AND TUTORIALS, and the Area Editor for Signal Processing: Image Communication. He was also the Guest Editor of a few special issues in international journals, including as the Guest Editor of the IEEE JOURNAL OF SELECTED TOPICS IN SIGNAL PROCESSING. He was an APSIPA Distinguished Lecturer from 2021 to 2022 and also an IEEE Distinguished Lecturer of Circuits and Systems Society from 2019 to 2020. He is the Chair of the IEEE ICME Steering Committee from 2024 to 2025, and the Chair of IEEE Chengdu Section.
\end{IEEEbiography}

\end{document}